\documentclass[12pt]{article}

\usepackage{placeins}
\usepackage{rotating}
\usepackage{amsmath, amssymb, amsfonts, latexsym, mathtools}
\usepackage{amsthm, mathrsfs, bm}

\usepackage{booktabs}
\usepackage{multirow}
\usepackage{makecell}
\usepackage{adjustbox}
\usepackage{tabularx}

\usepackage{graphicx}
\usepackage{subcaption}
\usepackage{float}

\usepackage[ruled, lined, linesnumbered, commentsnumbered, longend]{algorithm2e}
\SetKwProg{Pro}{procedure}{}{end\ procedure}
\usepackage[table,x11names]{xcolor}
\usepackage{url}
\usepackage{hyperref}

\usepackage{courier}
\usepackage{helvet}

\usepackage[utf8]{inputenc}
\usepackage[english]{babel}
\usepackage{setspace}
\usepackage{comment}
\usepackage{enumitem}

\usepackage{caption}

\usepackage[nohead]{geometry}
\usepackage[round]{natbib}

\usepackage{tikz}
\usetikzlibrary{shapes.geometric, arrows, positioning}

\usepackage{authblk}

\definecolor{RoyalBlue}{rgb}{0.25,0.41,0.88}
\definecolor{light-gray}{gray}{0.97}
\definecolor{qblack}{gray}{0.30}

\hypersetup{
    colorlinks=true,
    linkcolor=blue,
    urlcolor=blue,
    citecolor=blue
}

\theoremstyle{plain}   

\newtheorem{theorem}{Theorem}[section]
\newtheorem{proposition}[theorem]{Proposition}

\newtheorem{corollary}[theorem]{Corollary}

\theoremstyle{definition}  

\theoremstyle{remark}  

\numberwithin{equation}{section}

\title{\Large Classification-risk-optimal label acquisition}

\author[1]{Fariborz Setoudehtazangi}

\author[2]{Geoffrey J. McLachlan%
\thanks{Corresponding author.
Email:
\href{mailto:g.mclachlan@uq.edu.au}
{g.mclachlan@uq.edu.au}}}

\affil[1]{Dipartimento di Scienze Statistiche,
Università di Padova, Padova, Italy}

\affil[2]{School of Mathematics and Physics,
The University of Queensland, Australia}
\begin{document}

\date{}
\maketitle

\begin{abstract}
We study how a limited labeling budget should be allocated to minimize
multiclass zero--one classification risk. We consider parametric
classification problems in which features are observed for all sampling
units while class labels can be acquired selectively. By combining the
Fisher information supplied by an acquired label with the local geometry of
multiclass excess risk, we derive an acquisition criterion that minimizes the leading asymptotic coefficient of expected multiclass excess risk. The resulting rule values a label according to how strongly its
information is aligned with parameter directions that perturb the active
Bayes decision boundary, rather than according to posterior uncertainty or
global parameter information alone. We characterize the oracle acquisition
design, establish its threshold structure, and derive face-specific and
cost-sensitive extensions. An analytic example shows that posterior
uncertainty and classification value can produce different, and even
reversed, acquisition rankings. We further develop a two-stage adaptive
procedure that attains the oracle leading-risk criterion under regularity
conditions and provide explicit results for Gaussian discriminant analysis.
Three-class QDA experiments illustrate the resulting acquisition geometry,
while an application to the six-class Statlog Landsat Satellite data shows
that classification-risk acquisition can differ materially from both
uncertainty-based acquisition and the complete-classification-information
comparator. The adaptive classification-risk design attains lower mean error
than this Fisher comparator across the labeling budgets considered, although
it does not uniformly outperform entropy or margin sampling and differences
among the targeted strategies become small as the labeling budget increases.
\end{abstract}

\noindent\textbf{Keywords:}
active learning;
classification risk;
Fisher information;
label acquisition;
multiclass classification;
optimal design;
selective labeling.
\section{Introduction}
\label{sec:introduction}

In many classification problems, feature measurements are relatively
inexpensive, whereas obtaining class labels requires expert assessment,
laboratory testing, manual annotation, or another costly intervention. When
only a limited number of labels can be acquired, the statistical problem is
therefore not merely how to estimate a classifier from partially labeled
data, but which observations should be labeled. This is the central problem
of active learning: labels are acquired selectively so that a prescribed
learning or predictive objective can be achieved with a smaller labeling
budget
\citep{lewis1994sequential,lewis1994heterogeneous,settles2009active}.

A widely used strategy is uncertainty sampling, which queries observations for
which the current classifier is least confident. Least-confidence, entropy,
and margin-based rules implement this principle by concentrating labels where
posterior class probabilities are difficult to distinguish. These rules are
appealing because observations near an estimated decision boundary appear
particularly informative for classification. Posterior uncertainty at an
individual feature value, however, does not by itself characterize how
observing the corresponding label will affect estimation of the parameters
that determine the future decision boundary. Consequently, observations with
essentially identical posterior uncertainty can have substantially different
values for improving the eventual classifier. This limitation has motivated
acquisition criteria that account for an observation's effect on model fitting,
prediction, or risk, rather than relying on posterior ambiguity alone
\citep{bach2006active,schein2007active,imberg2020optimal,
bickfordsmith2023prediction}.

A second line of work formulates active learning through statistical
information and optimal experimental design. Early analyses related the value
of unlabeled observations to Fisher-information criteria
\citep{zhang2000probability}, while subsequent work developed sequential and
batch Fisher-information methods and provided asymptotic justification for
Fisher-information-ratio objectives
\citep{hoi2006batch,sourati2017asymptotic}. For multinomial logistic
regression, \citet{chen2023firal} established finite-sample connections between
the Fisher information ratio and excess risk and developed the FIRAL selection
algorithm. Related ideas arise in optimal subsampling, where inclusion
probabilities are chosen to control the asymptotic covariance of an estimator.
For logistic and softmax regression, for example, trace-based optimality
criteria produce sampling designs that combine a candidate observation's
information contribution with the covariance structure of the resulting
estimator \citep{wang2018optimal,yao2019optimal}. These developments establish
that label value depends not only on posterior uncertainty, but also on the
information supplied by an observation and the parameter directions in which
that information is supplied.

Acquisition has also been targeted more directly to predictive performance or
downstream decisions. Expected-error-reduction methods select a query by
averaging the predictive error anticipated after its possible labels have been
observed and the model updated
\citep{roy2001toward,mussmann2022active}. Asymptotic approaches have optimized
sampling distributions for generalization loss or prediction error in
generalized linear and pool-based models
\citep{bach2006active,imberg2020optimal}. More broadly, theoretical work on
active learning has directly studied classification performance under
zero--one loss, including the relationship between surrogate-loss optimization
and zero--one classification risk \citep{hanneke2019surrogate}. In the Bayesian
setting, prediction-oriented acquisition measures information about future
predictions rather than model parameters
\citep{bickfordsmith2023prediction}, while decision- and loss-oriented
procedures tailor acquisition to the ultimate decision problem or its
associated loss \citep{filstroff2024targeted,huang2026loss}. Thus neither the
general principle of targeting acquisition to predictive performance nor the
observation that uncertainty sampling can be suboptimal is new.

The objective developed here is distinguished by the particular predictive
loss from which it is derived and by the geometry that this loss induces.
Specifically, we consider population zero--one classification risk for a
regular parametric multiclass classifier. Unlike a smooth likelihood or
prediction loss, zero--one risk changes when perturbations of the model move
the decision boundary. In a multiclass problem, that boundary is composed of
portions of pairwise equality surfaces, and only the portions on which the
corresponding classes are jointly optimal affect the Bayes rule. We refer to
these portions as the \emph{active pairwise Bayes faces}. Consequently, the
classification relevance of a parameter direction is determined by how it
perturbs these active faces, not simply by how accurately that direction can
be estimated under the likelihood.

This boundary geometry was characterized in our preceding work on informative
label missingness \citep{SetoudehtazangiMcLachlan2026}. For a regular
parametric multiclass classifier and a local perturbation \(h\) from the
data-generating parameter \(\theta_0\), the population risk satisfies

$$
R(\theta_0+h)-R^*
=
\frac{1}{2}h^\top H_R h+o(\lVert h\rVert^2),
$$

where

$$
H_R=\sum_{k<l}H_{kl}.
$$

Each matrix \(H_{kl}\) is obtained by integrating the sensitivity of the
pairwise decision boundary over the active Bayes face separating classes
\(k\) and \(l\). Hence \(H_R\) is not an externally chosen weighting matrix:
it is the local curvature of multiclass population zero--one risk. That work
used \(H_R\) to determine how information gains and losses induced by an
informative label-missingness mechanism affect classification efficiency.
Related work has studied the information carried by uncertainty-dependent
missing labels and its implications for semi-supervised classification
\citep{wang2026learning}. In those problems, the label-availability mechanism
is externally specified. Here we instead treat label acquisition as a
controlled design variable and ask how a limited labeling budget should be
allocated to minimize classification risk.

We consider a parametric multiclass model in which the feature vector \(Y\) is
observed for every sampling unit and the corresponding class label \(Z\) is
acquired with investigator-controlled probability \(a(Y)\). The Fisher
information under an acquisition design \(a\) is

$$
I(a)
=
I_Y+
E\!\left\{a(Y)I_{Z\mid Y}(Y)\right\},
$$

where \(I_Y\) denotes the information contained in the marginal distribution
of the features and \(I_{Z\mid Y}(y)\) is the conditional information supplied
by observing the label at \(Y=y\). Combining this information identity with
the local expansion of zero--one risk yields

$$
E\{R(\widehat{\theta}_a)\}-R^*
=
\frac{1}{2n}
\operatorname{tr}\!\left\{H_R I(a)^{-1}\right\}
+o(n^{-1}).
$$

The leading expected excess-risk coefficient therefore depends on the
interaction between the information generated by the acquisition design and
the parameter directions that locally affect the multiclass decision
boundary.

Subject to the labeling-budget constraint \(E\{a(Y)\}\leq\rho\), this leads to
the oracle design problem

$$
\min_{a}\;
\operatorname{tr}\!\left\{H_R I(a)^{-1}\right\}.
$$

For fixed \(H_R\), this is an \(L\)-optimal, equivalently weighted
\(A\)-optimal, trace-inverse criterion
\citep{pukelsheim2006optimal}. Its distinctive feature is not the generic
trace-inverse form, but the derivation of its target matrix from population
zero--one risk. Unlike likelihood- or test-information-oriented targets,
\(H_R\) weights parameter directions according to their local contribution to
movement of the active Bayes boundary.

The marginal classification value of acquiring the label at feature value
\(y\) is

$$
\psi_a(y)
=
\operatorname{tr}\!\left[
H_R I(a)^{-1}
I_{Z\mid Y}(y)
I(a)^{-1}
\right].
$$

This expression combines three distinct components: the conditional
information contributed by the candidate label, the information already
available under the current design, and the zero--one-risk importance of the
affected parameter directions. The oracle optimality conditions have a
threshold form: labels are acquired where the marginal classification value
exceeds a budget-determined threshold, subject to possible randomization on
the threshold set. Moreover, the representation
\(H_R=\sum_{k<l}H_{kl}\) decomposes \(\psi_a(y)\) into face-specific
contributions. An observation need not itself lie on a particular Bayes face
to be valuable for that face; its label may improve estimation of parameters
that determine the face elsewhere in the feature space.

This characterization clarifies the distinction from both uncertainty
sampling and the Fisher comparator considered in this paper. We construct an
analytic example in which observations with identical posterior uncertainty
have different classification values and, more strongly, in which uncertainty
and classification-risk acquisition produce opposite rankings. We also
compare the proposed target \(H_R\) with the complete-classification
information target \(I_{\mathrm{CC}}\). Both yield trace-inverse
information-based designs; the difference lies in the directions emphasized
by their target matrices. The relevant contrast is therefore not
``uncertainty versus information'', but posterior ambiguity versus the
directional value of label information, and, within information-based design,
likelihood-oriented weighting versus weighting induced by local zero--one
classification risk.

Because the oracle design depends on the unknown data-generating parameter, we
develop a two-stage adaptive procedure. A pilot sample is used to estimate the
classification model, the relevant information matrices, and the curvature
of the active Bayes boundary; the remaining labeling budget is then allocated
using the resulting plug-in acquisition rule. Under the stated regularity,
consistency, and uniform convergence conditions, the plug-in design attains
the oracle leading-risk coefficient asymptotically. The framework also permits
nonconstant labeling costs, for which the relevant marginal criterion is
classification value per unit cost.

The numerical investigation considers three-class Gaussian quadratic
discriminant analysis. Unequal class covariance matrices generate curved
pairwise Bayes boundaries and make the relationship between acquisition
locations and active-face geometry directly visible. We compare random
acquisition, entropy and margin sampling, Fisher-information design, the
adaptive classification-risk design, and its oracle counterpart. A
heterogeneous QDA configuration examines the setting in which the
classification-risk and information targets differ substantially, while a
near-linear control configuration examines a regime in which their distinction
is weaker. The simulations show how classification-risk weighting changes the
locations selected for labeling and indicate that its consequences are most
pronounced under small labeling budgets.

We additionally analyze the six-class Statlog Landsat Satellite data
\citep{Srinivasan1993}, using the supplied training--test split and repeated
pilot-and-acquisition experiments. The application provides a direct
comparison of posterior uncertainty, complete-classification-information
targeting, and classification-risk targeting in a higher-dimensional
multiclass problem. A seven-class application to the Dry Bean data
\citep{KokluOzkan2020} is reported in the Supplementary Material.

The contribution of this paper is therefore not a new generic trace-optimal
design criterion, the general idea of prediction- or loss-targeted
acquisition, or the observation that uncertainty sampling may be inefficient.
Rather, it is the derivation of a label-acquisition criterion from the local
second-order geometry of multiclass population zero--one risk, the
decomposition of its target matrix over active pairwise Bayes faces, and the
resulting allocation rule that couples these classification-relevant
directions to the conditional information supplied by an acquired label. This
construction provides a classification-specific interpretation of label value
and identifies precisely when two observations having similar posterior
uncertainty can differ in their value for improving the final classifier.

The remainder of the paper is organized as follows.
Section~\ref{sec:acquisition} formulates the controlled label-acquisition
model, derives its Fisher-information decomposition, and connects the
resulting information matrix to the leading expected multiclass excess risk.
Section~\ref{sec:optimal-design} introduces the classification-risk-optimal
oracle design, establishes its convexity, and derives the marginal-value and
threshold characterizations. Section~\ref{sec:uncertainty} compares
classification value with posterior uncertainty and provides an analytic
example in which the two criteria produce different, and potentially
opposite, acquisition rankings. Section~\ref{sec:face-specific-value}
develops the active-face decomposition and extends the framework to
nonconstant labeling costs. Section~\ref{sec:adaptive} constructs the
two-stage plug-in procedure and establishes its first-order oracle
equivalence. Section~\ref{sec:gaussian-specialization} specializes the
general theory to Gaussian linear and quadratic discriminant analysis.
Sections~\ref{sec:numerical} and~\ref{sec:realdata} present the simulation
study and the Statlog Landsat Satellite application, respectively.
Section~\ref{sec:discussion} discusses the interpretation, scope, and
limitations of the proposed framework, and
Section~\ref{sec:conclusion} concludes.

\section{Controlled label acquisition and classification risk}
\label{sec:acquisition}

Let \(Y\in\mathcal{Y}\subseteq\mathbb{R}^{p}\) denote an observed feature
vector and let \(Z\in\{1,\ldots,g\}\) denote its class label. We consider a
regular parametric classification model
\[
p_{\theta}(y,z)
=
p_{\theta}(y)\tau_{z}(y;\theta),
\qquad
\tau_k(y;\theta)
=
\Pr_{\theta}(Z=k\mid Y=y),
\]
where \(\theta\in\Theta\subseteq\mathbb{R}^{d}\) is identifiable and
\(\Theta\) is open.

The feature vector is assumed to be available for every sampling unit, whereas
the class label is acquired selectively. Let
\[
\Delta=
\begin{cases}
1, & \text{if the class label is acquired},\\
0, & \text{otherwise},
\end{cases}
\]
and suppose that, conditional on \(Y=y\),
\begin{equation}
\Pr(\Delta=1\mid Y=y,Z)=a(y),
\qquad 0\leq a(y)\leq 1.
\label{eq:acquisition-mechanism}
\end{equation}
The measurable function \(a:\mathcal{Y}\to[0,1]\) is chosen by the
investigator and will be referred to as the \emph{acquisition design}.
Importantly, \(a\) is regarded throughout this section as fixed and known and
does not depend on the unknown parameter \(\theta\). Thus the acquisition
indicator records a design decision rather than an additional
parameter-dependent source of information.

The observed datum is
\[
O=(Y,\Delta,\Delta Z).
\]
Under \eqref{eq:acquisition-mechanism}, its likelihood contribution for
\(\theta\), up to factors that do not depend on \(\theta\), is
\begin{equation}
L_a(\theta;O)
\propto
p_{\theta}(Y)
\{\tau_Z(Y;\theta)\}^{\Delta}.
\label{eq:observed-likelihood-acquisition}
\end{equation}
Write
\[
S_Y(Y;\theta)=\nabla_{\theta}\log p_{\theta}(Y)
\]
for the feature score and
\[
S_{Z\mid Y}(Y,Z;\theta)
=
\nabla_{\theta}\log \tau_Z(Y;\theta)
\]
for the conditional class-label score. Define
\[
I_Y(\theta)
=
E_{\theta}
\left[
S_Y S_Y^{\top}
\right]
\]
and
\begin{equation}
I_{Z\mid Y}(\theta;y)
=
E_{\theta}
\left[
S_{Z\mid Y}S_{Z\mid Y}^{\top}
\mid Y=y
\right].
\label{eq:conditional-label-information}
\end{equation}
The matrix \(I_{Z\mid Y}(\theta;y)\) is the information supplied by observing
the class label once the feature value \(y\) is known.

The following result gives the information available under a fixed
acquisition design.

\begin{proposition}
\label{prop:design-information}
Assume the usual regularity conditions for Fisher-information calculations.
Under the controlled acquisition mechanism
\eqref{eq:acquisition-mechanism}, the Fisher information for \(\theta\) is
\begin{equation}
I(a;\theta)
=
I_Y(\theta)
+
E_{\theta}
\left[
a(Y)I_{Z\mid Y}(\theta;Y)
\right].
\label{eq:design-information}
\end{equation}
Equivalently, if
\[
I_{\mathrm{CC}}(\theta)
=
I_Y(\theta)
+
E_{\theta}
\left[
I_{Z\mid Y}(\theta;Y)
\right]
\]
denotes the information under complete classification, then
\begin{equation}
I(a;\theta)
=
I_{\mathrm{CC}}(\theta)
-
E_{\theta}
\left[
\{1-a(Y)\}I_{Z\mid Y}(\theta;Y)
\right].
\label{eq:design-information-loss}
\end{equation}
\end{proposition}

The proof is given in Supplementary Section~\ref{sec:supp-acquisition}.

Proposition~\ref{prop:design-information} separates the contribution of the
always-observed features from the contribution of selectively acquired class
labels. In contrast to informative label missingness generated by a
parameter-dependent mechanism, the acquisition indicator itself contributes
no Fisher information about \(\theta\): its distribution is fixed by design.
The statistical role of \(a(y)\) is instead to determine where in feature
space the conditional label information
\(I_{Z\mid Y}(\theta;y)\) is retained.

We now connect this information structure to classification performance.
For \(k=1,\ldots,g\), let
\[
r_k(y;\theta)=\Pr_{\theta}(Z=k)p_{\theta}(y\mid Z=k)
\]
denote the prior-weighted class density, and let
\[
C_{\theta}(y)=\arg\max_{1\leq k\leq g}r_k(y;\theta)
\]
be the associated Bayes rule, with ties resolved by a fixed deterministic
rule. Under the boundary regularity conditions used below, the set of ties
has probability zero under the true distribution and therefore does not
affect the classification risk. Classification risk is evaluated under the
true parameter \(\theta_0\):
\[
R(\theta)
=
\Pr_{\theta_0}\{C_{\theta}(Y)\neq Z\},
\qquad
R^*=R(\theta_0).
\]

For regular multiclass Bayes boundaries, the local excess risk admits the
quadratic expansion
\begin{equation}
R(\theta_0+h)-R^*
=
\frac{1}{2}h^{\top}H_Rh
+
o(\|h\|^2),
\qquad h\to0,
\label{eq:risk-quadratic}
\end{equation}
where \(H_R\succeq0\) is the classification-risk curvature matrix. In
particular, if \(\mathcal{F}_{kl}\) denotes the active Bayes face separating
classes \(k\) and \(l\), then
\begin{equation}
H_R
=
\sum_{1\leq k<l\leq g}H_{kl},
\qquad
H_{kl}
=
\int_{\mathcal{F}_{kl}}
\frac{
b_{kl}(s)b_{kl}(s)^{\top}
}{
\|\nabla_y g_{kl}(s;\theta_0)\|
}
\,dS(s),
\label{eq:risk-curvature}
\end{equation}
with
\[
g_{kl}(y;\theta)
=
r_k(y;\theta)-r_l(y;\theta),
\qquad
b_{kl}(s)
=
\nabla_{\theta}g_{kl}(s;\theta_0).
\]
Thus \(H_R\) weights parameter directions according to their first-order
effect on the active multiclass decision boundary.

The consequence for label acquisition is immediate.

\begin{proposition}
\label{prop:design-risk}
Let \(a\) be a fixed acquisition design for which \(I(a;\theta_0)\) is
positive definite, and let \(\widehat{\theta}_a\) be an efficient regular
estimator satisfying
\[
\sqrt{n}
(\widehat{\theta}_a-\theta_0)
\overset{d}{\longrightarrow}
N\!\left(
0,
I(a;\theta_0)^{-1}
\right).
\]
Assume also the boundary regularity conditions required for
\eqref{eq:risk-quadratic} and sufficient moment conditions ensuring uniform
integrability of
\(n\{R(\widehat{\theta}_a)-R^*\}\). Then
\begin{equation}
E_{\theta_0}
\left[
R(\widehat{\theta}_a)
\right]
-
R^*
=
\frac{1}{2n}
\operatorname{tr}
\left\{
H_R I(a;\theta_0)^{-1}
\right\}
+
o(n^{-1}).
\label{eq:design-leading-risk}
\end{equation}
\end{proposition}

The proof is given in Supplementary Section~\ref{sec:supp-acquisition}.

Equation~\eqref{eq:design-leading-risk} is the starting point for the design
problem studied in the remainder of the paper. It shows that a label is not
valuable merely because its class membership is uncertain. Its value depends
on whether the information supplied by that label reduces estimation
uncertainty in parameter directions that matter for the active Bayes
boundary. Consequently, the relevant acquisition criterion is not Fisher
information alone, nor posterior uncertainty alone, but their interaction
through the classification-risk geometry \(H_R\).

In the next section, we use \eqref{eq:design-leading-risk} to formulate and
characterize the label-acquisition design that minimizes the leading
multiclass excess classification risk under a labeling budget.
\section{Classification-risk-optimal label acquisition}
\label{sec:optimal-design}

We now determine how a limited labeling budget should be allocated over the
feature space. Throughout this section, all information quantities are
evaluated at the true parameter value \(\theta_0\). To simplify notation,
write
\[
J(y)
=
I_{Z\mid Y}(\theta_0;y),
\qquad
I(a)
=
I_Y(\theta_0)+E_{\theta_0}\{a(Y)J(Y)\},
\]
and let \(H_R\) denote the classification-risk curvature matrix introduced in
Section~\ref{sec:acquisition}.

Suppose that labels may be acquired for a proportion at most
\(\rho\in(0,1)\) of the population. The admissible acquisition designs are
measurable functions \(a:\mathcal{Y}\to[0,1]\) satisfying
\[
E_{\theta_0}\{a(Y)\}\leq\rho.
\]
By Proposition~\ref{prop:design-risk}, the leading excess classification
risk under design \(a\) is proportional to
\[
\Phi(a)
=
\operatorname{tr}\{H_R I(a)^{-1}\}.
\]
The oracle acquisition problem is therefore
\begin{equation}
\inf_{0\leq a\leq1,\;E\{a(Y)\}\leq\rho}
\Phi(a).
\label{eq:oracle-design-inequality}
\end{equation}

Since \(J(y)\succeq0\), increasing the probability of acquiring labels can
only increase \(I(a)\) in the Loewner ordering and hence cannot increase
\(\Phi(a)\). Consequently, the optimal value in
\eqref{eq:oracle-design-inequality} is unchanged if the budget constraint is
replaced by equality. We shall therefore work with
\begin{equation}
\mathcal A_\rho
=
\left\{
a:\mathcal Y\rightarrow[0,1]:
a\ \text{measurable},\
E_{\theta_0}\{a(Y)\}=\rho
\right\}.
\label{eq:design-class}
\end{equation}
The oracle design problem becomes
\begin{equation}
a^\star
\in
\arg\min_{a\in\mathcal A_\rho}
\Phi(a).
\label{eq:oracle-design}
\end{equation}

The next result characterizes this problem completely at the population
level. For a design \(a\), define
\begin{equation}
\psi_a(y)
=
\operatorname{tr}
\left\{
H_R I(a)^{-1}
J(y)
I(a)^{-1}
\right\}.
\label{eq:classification-value}
\end{equation}
We call \(\psi_a(y)\) the \emph{classification value} of acquiring the class
label at feature value \(y\). It measures the marginal reduction in the
leading classification-risk criterion generated by allocating additional
labeling probability at \(y\).

\begin{theorem}
\label{thm:oracle-design}
Suppose that \(J(Y)\) is integrable, \(H_R\succeq0\), and the information
matrices \(I(a)\) are uniformly nonsingular over
\(a\in\mathcal A_\rho\). Then:

\begin{enumerate}
\item[(i)]
The criterion
\[
\Phi(a)=\operatorname{tr}\{H_RI(a)^{-1}\}
\]
is convex on \(\mathcal A_\rho\), and the oracle problem
\eqref{eq:oracle-design} admits at least one solution.

\item[(ii)]
For every bounded measurable perturbation \(h\) such that
\(a+th\) remains admissible for all sufficiently small \(t\geq0\),
the right directional derivative of \(\Phi\) at \(a\) is
\begin{equation}
D\Phi(a)[h]
=
-
E_{\theta_0}
\left[
h(Y)\psi_a(Y)
\right].
\label{eq:directional-derivative}
\end{equation}

\item[(iii)]
A design \(a^\star\in\mathcal A_\rho\) is oracle optimal if and only if
\begin{equation}
E_{\theta_0}
\left[
a(Y)\psi_{a^\star}(Y)
\right]
\leq
E_{\theta_0}
\left[
a^\star(Y)\psi_{a^\star}(Y)
\right]
\qquad
\text{for every }a\in\mathcal A_\rho.
\label{eq:equivalence-condition}
\end{equation}

\item[(iv)]
Equivalently, there exists a constant
\(\lambda^\star\in\mathbb R\) such that, almost surely,
\begin{equation}
\begin{cases}
\psi_{a^\star}(Y)\leq\lambda^\star,
&
a^\star(Y)=0,
\\[3pt]
\psi_{a^\star}(Y)=\lambda^\star,
&
0<a^\star(Y)<1,
\\[3pt]
\psi_{a^\star}(Y)\geq\lambda^\star,
&
a^\star(Y)=1.
\end{cases}
\label{eq:threshold-characterization}
\end{equation}
\end{enumerate}
\end{theorem}

The proof is given in Supplementary Section~\ref{sec:supp-oracle-design}. The uniform nonsingularity
assumption excludes acquisition designs for which one or more parameter
directions required for regular estimation are unidentified. This condition
is automatic when \(I_Y\) is positive definite, but it can be substantive
when the feature distribution alone does not identify all components of
\(\theta\). In that case, the acquired-label information must provide
sufficient information in the otherwise unidentified directions.

Theorem~\ref{thm:oracle-design} provides a general equivalence
characterization of classification-risk-optimal label acquisition. Although
\(\psi_a(y)\) depends on the design through \(I(a)\), at the optimum the
problem has a threshold structure: labels are acquired where their marginal
classification value is largest. Randomized acquisition can occur only on a
level set of \(\psi_{a^\star}\).

In particular, if the distribution of
\(\psi_{a^\star}(Y)\) has no atom at the threshold
\(\lambda^\star\), then the oracle design is deterministic almost surely.

\begin{corollary}
\label{cor:bang-bang-design}
Under the conditions of Theorem~\ref{thm:oracle-design}, suppose additionally
that
\[
\Pr_{\theta_0}
\{\psi_{a^\star}(Y)=\lambda^\star\}=0.
\]
Then
\begin{equation}
a^\star(y)
=
\mathbf 1
\{\psi_{a^\star}(y)>\lambda^\star\}
\quad\text{for almost every }y,
\label{eq:bang-bang-design}
\end{equation}
where \(\lambda^\star\) is chosen so that
\[
\Pr_{\theta_0}
\{\psi_{a^\star}(Y)>\lambda^\star\}
=
\rho.
\]
\end{corollary}

The characterization is implicit because the value function
\(\psi_{a^\star}\) itself depends on the information generated by the optimal
design. This dependence is important. The oracle rule does not rank
observations using a fixed pointwise measure of class ambiguity. Rather,
\(\psi_a(y)\) combines three distinct quantities:
\[
J(y),
\qquad
I(a)^{-1},
\qquad
H_R.
\]
The first describes the conditional information supplied by the label at
\(y\); the second accounts for the directions in parameter space that remain
poorly estimated under the current acquisition design; and the third retains
only those parameter directions that affect classification risk through the
active Bayes boundary.

This interpretation can be made more explicit. Let
\[
G_a
=
I(a)^{-1}H_RI(a)^{-1}.
\]
By cyclic invariance of the trace,
\begin{equation}
\psi_a(y)
=
\operatorname{tr}\{G_aJ(y)\}.
\label{eq:value-alignment}
\end{equation}
Thus the value of acquiring a label is determined not by the magnitude of
\(J(y)\) alone, but by its alignment with the classification-relevant
estimation directions encoded by \(G_a\). An observation may therefore be
highly informative about the model while contributing little to the
parameter directions that move an active Bayes face. Conversely, a label
with more modest total information can have high classification value when
its information is concentrated in precisely those directions.

The same formulation also makes the distinction from global
Fisher-information design explicit. Let
\[
I_{\mathrm{CC}}
=
I_Y+E\{J(Y)\}
\]
denote the complete-classification information and consider the
information-targeted criterion
\begin{equation}
\Phi_{\mathrm F}(a)
=
\operatorname{tr}
\left\{
I_{\mathrm{CC}}I(a)^{-1}
\right\}.
\label{eq:fisher-design-criterion}
\end{equation}
Its marginal value is
\begin{equation}
\psi_{\mathrm F,a}(y)
=
\operatorname{tr}
\left\{
I_{\mathrm{CC}}
I(a)^{-1}
J(y)
I(a)^{-1}
\right\}.
\label{eq:fisher-classification-comparison}
\end{equation}
Thus the complete-classification-information and classification-risk
criteria have the same trace-inverse design structure but different target
matrices: \(I_{\mathrm{CC}}\) for likelihood-oriented estimation relative
to the complete-classification benchmark and \(H_R\) for local zero--one
classification risk. The former weights directions according to
\(I_{\mathrm{CC}}\), whereas the latter weights them according to their
local effects on the active Bayes boundary. This comparison concerns the
particular \(I_{\mathrm{CC}}\)-targeted Fisher criterion considered here;
it is not intended to characterize all Fisher-information-based
active-learning methods. The proposed contribution
therefore lies not in the generic trace-inverse optimization itself, but in
the classification-risk target \(H_R\) obtained from the local geometry of
multiclass zero--one loss.

The next section develops the complementary distinction from
posterior-uncertainty-based acquisition.
\section{Classification value and posterior uncertainty}
\label{sec:uncertainty}

A common principle in label acquisition is to prioritize observations for
which the current classifier is most uncertain. For a multiclass posterior
vector
\[
\boldsymbol{\tau}(y)
=
\{\tau_1(y),\ldots,\tau_g(y)\}^{\top},
\]
typical uncertainty scores depend only on
\(\boldsymbol{\tau}(y)\). Examples include posterior entropy,
\begin{equation}
U_{\mathrm{ent}}(y)
=
-\sum_{k=1}^{g}
\tau_k(y)\log\tau_k(y),
\label{eq:entropy-score}
\end{equation}
and the posterior margin between the two largest class probabilities.

The classification value in
\eqref{eq:classification-value} has a fundamentally different structure:
\[
\psi_a(y)
=
\operatorname{tr}
\left\{
H_R I(a)^{-1}
J(y)
I(a)^{-1}
\right\}.
\]
It depends not only on the posterior ambiguity at \(y\), but also on the
matrix-valued information supplied by observing the label and on the
alignment of that information with parameter directions that affect the
active Bayes boundary. Consequently, posterior uncertainty alone is not, in
general, sufficient to rank observations according to their value for
classification.

The distinction can be established explicitly in a simple regular model.
Let \(Y=(Y_1,Y_2)^{\top}\sim N_2(0,I_2)\), independently of the parameter,
and suppose that
\begin{equation}
\Pr_{\boldsymbol\beta}(Z=1\mid Y=y)
=
\sigma(\boldsymbol x^{\top}\boldsymbol\beta),
\qquad
\boldsymbol x=(1,y_1,y_2)^{\top},
\label{eq:logistic-example}
\end{equation}
where
\[
\sigma(t)=\frac{e^t}{1+e^t}.
\]
Consider the true parameter
\begin{equation}
\boldsymbol\beta_0=(0,b,0)^{\top},
\qquad b>0.
\label{eq:logistic-true-parameter}
\end{equation}
The Bayes boundary is then the line \(y_1=0\).

Suppose initially that labels are acquired uniformly,
\[
a_{\rho}(y)\equiv\rho,
\qquad 0<\rho<1.
\]
The following result gives both the uncertainty score and the exact
classification-value score in this model.

\begin{proposition}
\label{prop:uncertainty-counterexample}
Under \eqref{eq:logistic-example}--\eqref{eq:logistic-true-parameter}, let
\[
v(t)=\sigma(t)\{1-\sigma(t)\}
\]
and
\[
m_0=E\{v(bY_1)\}.
\]
For the uniform acquisition design \(a_{\rho}\),

\begin{equation}
\psi_{a_{\rho}}(y)
=
C_{\rho,b}\,
v(by_1)(1+y_2^2),
\qquad
C_{\rho,b}
=
\frac{\phi(0)}
{2b\rho^2m_0^2}
>0,
\label{eq:logistic-classification-value}
\end{equation}
where \(\phi\) denotes the standard normal density.

By contrast, posterior entropy satisfies
\begin{equation}
U_{\mathrm{ent}}(y)
=
h\{\sigma(by_1)\},
\label{eq:logistic-entropy}
\end{equation}
where
\[
h(u)=-u\log u-(1-u)\log(1-u).
\]
Thus entropy depends only on \(y_1\), whereas
\(\psi_{a_{\rho}}(y)\) depends additionally on \(y_2^2\).

In particular:

\begin{enumerate}
\item[(i)]
for any fixed \(y_1\) and any \(y_2\neq\widetilde y_2\) satisfying
\(y_2^2\neq\widetilde y_2^2\),
\[
U_{\mathrm{ent}}(y_1,y_2)
=
U_{\mathrm{ent}}(y_1,\widetilde y_2),
\]
while
\[
\psi_{a_{\rho}}(y_1,y_2)
\neq
\psi_{a_{\rho}}(y_1,\widetilde y_2);
\]

\item[(ii)]
there exist \(y^{(1)},y^{(2)}\in\mathbb R^2\) such that
\begin{equation}
U_{\mathrm{ent}}\{y^{(1)}\}
>
U_{\mathrm{ent}}\{y^{(2)}\},
\qquad
\psi_{a_{\rho}}\{y^{(1)}\}
<
\psi_{a_{\rho}}\{y^{(2)}\}.
\label{eq:ranking-reversal}
\end{equation}
\end{enumerate}
\end{proposition}

The proof is given in Supplementary Section~\ref{sec:supp-uncertainty}.

Proposition~\ref{prop:uncertainty-counterexample} shows that the distinction
between uncertainty and classification value is not merely a consequence of
multiclass complexity or nonlinear decision boundaries. It occurs already in
a regular binary logistic model with a linear Bayes boundary.

The geometry of the example is informative. Since
\[
\Pr(Z=1\mid Y=y)=\sigma(by_1),
\]
posterior uncertainty is determined entirely by the coordinate normal to the
Bayes boundary. Points having the same value of \(y_1\) therefore receive the
same entropy score regardless of their location along the boundary.
Nevertheless, perturbations of the coefficient of \(Y_2\) rotate the
decision boundary, and observations with large \(|y_2|\) carry greater
information about that classification-relevant direction. This contribution
appears through the factor
\[
1+y_2^2
\]
in \eqref{eq:logistic-classification-value}, but is invisible to posterior
entropy.

The same conclusion applies to binary margin-based uncertainty scores,
because these too are functions only of the posterior probability
\(\sigma(by_1)\). Hence no acquisition rule that ranks observations solely by
posterior uncertainty can, in general, reproduce the
classification-risk-optimal ranking.

There is nevertheless a useful special case in which the two principles can
agree. If the conditional label-information matrices satisfy
\begin{equation}
J(y)=c\{\boldsymbol{\tau}(y)\}J_0
\label{eq:uncertainty-agreement-condition}
\end{equation}
for a fixed positive-semidefinite matrix \(J_0\) and a scalar function \(c\),
then
\[
\psi_a(y)
=
c\{\boldsymbol{\tau}(y)\}
\operatorname{tr}
\left\{
H_RI(a)^{-1}J_0I(a)^{-1}
\right\}.
\]
In this restricted situation, classification value is ordered entirely by
the scalar \(c\{\boldsymbol{\tau}(y)\}\). If that scalar is itself monotone in
a chosen posterior-uncertainty measure, the two rankings coincide.

Outside such structurally restricted settings, however, the matrix
orientation of \(J(y)\) matters. The oracle criterion therefore distinguishes
between being uncertain about a label and obtaining a label that is useful
for reducing classification risk. The next section develops this distinction
further in the multiclass setting by decomposing classification value over
the active pairwise Bayes faces.

\section{Face-specific classification value and acquisition costs}
\label{sec:face-specific-value}

The classification-risk curvature has the additive representation
\[
H_R
=
\sum_{1\leq k<l\leq g}H_{kl},
\]
where \(H_{kl}\succeq0\) is contributed by the active Bayes face
\(\mathcal F_{kl}\) separating classes \(k\) and \(l\). This structure allows
the value of a candidate label to be resolved according to the parts of the
multiclass decision boundary whose estimation it improves.

For each active class pair \(k<l\), define
\begin{equation}
\Phi_{kl}(a)
=
\operatorname{tr}
\left\{
H_{kl}I(a)^{-1}
\right\},
\label{eq:face-risk-criterion}
\end{equation}
and
\begin{equation}
\psi_{kl,a}(y)
=
\operatorname{tr}
\left\{
H_{kl}I(a)^{-1}
J(y)
I(a)^{-1}
\right\}.
\label{eq:face-classification-value}
\end{equation}
The quantity \(\Phi_{kl}(a)/(2n)\) is the leading expected excess-risk
contribution associated with the active face \(\mathcal F_{kl}\), whereas
\(\psi_{kl,a}(y)\) measures the marginal value of acquiring the label at
\(y\) for reducing that contribution.

\begin{proposition}
\label{prop:face-decomposition}
Under the conditions of Theorem~\ref{thm:oracle-design},

\begin{equation}
\Phi(a)
=
\sum_{1\leq k<l\leq g}
\Phi_{kl}(a),
\label{eq:criterion-face-decomposition}
\end{equation}
and
\begin{equation}
\psi_a(y)
=
\sum_{1\leq k<l\leq g}
\psi_{kl,a}(y).
\label{eq:value-face-decomposition}
\end{equation}

Moreover,
\[
\Phi_{kl}(a)\geq0,
\qquad
\psi_{kl,a}(y)\geq0,
\]
and for every admissible perturbation \(h\),
\begin{equation}
D\Phi_{kl}(a)[h]
=
-
E_{\theta_0}
\left[
h(Y)\psi_{kl,a}(Y)
\right].
\label{eq:face-directional-derivative}
\end{equation}
\end{proposition}

The proof is given in Supplementary Section~\ref{sec:supp-face-specific}.

Proposition~\ref{prop:face-decomposition} yields a direct multiclass
interpretation of the acquisition criterion. A label may be valuable because
the information it supplies is aligned strongly with parameter directions
that move one particular active Bayes face, with several faces
simultaneously, or with none of them to an appreciable extent. In particular,
a large amount of total parameter information does not imply a large
classification value if that information is concentrated in directions that
have little effect on the active decision boundary.

It is important that
\(\psi_{kl,a}(y)\) not be interpreted as assigning the observation \(y\) to
the class pair \((k,l)\). The observation need not lie on, or even near,
\(\mathcal F_{kl}\). Rather,
\(\psi_{kl,a}(y)\) quantifies how the conditional information obtained from
its label propagates through parameter estimation to the leading
classification risk associated with that face. The relationship is therefore
inferential rather than spatial.

The decomposition also permits the contribution of individual faces to the
overall acquisition value to be examined through
\begin{equation}
\omega_{kl,a}(y)
=
\frac{\psi_{kl,a}(y)}
{\psi_a(y)},
\qquad
\psi_a(y)>0,
\label{eq:face-value-share}
\end{equation}
for which
\[
\omega_{kl,a}(y)\geq0,
\qquad
\sum_{k<l}\omega_{kl,a}(y)=1.
\]
These quantities provide a diagnostic for identifying which parts of a
multiclass decision boundary contribute most strongly to the acquisition
value of particular observations.

So far, every acquired label has been assigned the same cost. In many
applications, however, obtaining the class label can require different
amounts of effort or expense for different observations. Let
\[
c:\mathcal Y\rightarrow(0,\infty)
\]
denote a known acquisition-cost function and suppose that the average
available budget is \(B\). The corresponding oracle problem is
\begin{equation}
\min_{0\leq a\leq1}
\Phi(a)
\qquad
\text{subject to}
\qquad
E_{\theta_0}\{c(Y)a(Y)\}\leq B.
\label{eq:cost-oracle-problem}
\end{equation}

The relevant quantity is then the classification value obtained per unit
cost,
\begin{equation}
\chi_a(y)
=
\frac{\psi_a(y)}{c(y)}.
\label{eq:value-per-cost}
\end{equation}

\begin{corollary}
\label{cor:cost-sensitive-design}
Suppose that \(J(Y)\) is integrable, \(H_R\succeq0\), and the information
matrices \(I(a)\) are uniformly nonsingular over the cost-feasible designs
satisfying
\[
0\leq a\leq1,
\qquad
E_{\theta_0}\{c(Y)a(Y)\}\leq B.
\]
Assume in addition that
\[
0<c_{\min}\leq c(Y)\leq c_{\max}<\infty
\]
almost surely, with
\[
0<B<E_{\theta_0}\{c(Y)\}.
\]
Then the optimal value of \eqref{eq:cost-oracle-problem} is attained by a
design that exhausts the available budget. For any such oracle design
\(a_B^\star\), there exists a constant
\(\lambda_B^\star\) such that, almost surely,
\begin{equation}
\begin{cases}
\chi_{a_B^\star}(Y)\leq\lambda_B^\star,
&
a_B^\star(Y)=0,
\\[3pt]
\chi_{a_B^\star}(Y)=\lambda_B^\star,
&
0<a_B^\star(Y)<1,
\\[3pt]
\chi_{a_B^\star}(Y)\geq\lambda_B^\star,
&
a_B^\star(Y)=1.
\end{cases}
\label{eq:cost-threshold-rule}
\end{equation}
If
\[
\Pr_{\theta_0}
\{\chi_{a_B^\star}(Y)=\lambda_B^\star\}=0,
\]
then
\begin{equation}
a_B^\star(y)
=
\mathbf 1
\{
\chi_{a_B^\star}(y)>\lambda_B^\star
\}
\quad\text{almost surely},
\label{eq:cost-bang-bang}
\end{equation}
with the threshold chosen so that
\[
E_{\theta_0}
\left[
c(Y)
\mathbf 1
\{\chi_{a_B^\star}(Y)>\lambda_B^\star\}
\right]
=
B.
\]
\end{corollary}

The proof is given in Supplementary Section~\ref{sec:supp-face-specific}.

Thus heterogeneous labeling costs do not alter the underlying
classification-risk principle. They change the ordering criterion from the
absolute marginal value
\[
\psi_a(y)
\]
to the marginal value per unit acquisition cost
\[
\frac{\psi_a(y)}{c(y)}.
\]
Consequently, an expensive observation can be passed over in favor of a less
informative but substantially cheaper label if the latter produces a greater
reduction in leading classification risk per unit cost.

The oracle characterizations in this and the preceding section depend on the
unknown data-generating parameter through \(I(a)\), \(J(y)\), and \(H_R\).
They therefore describe the target design rather than an immediately
implementable procedure. The next section addresses this issue by replacing
the population quantities with sequentially updated estimates and studying
whether the resulting adaptive acquisition rule approaches the oracle
classification-risk criterion.

\section{Adaptive classification-risk-optimal acquisition}
\label{sec:adaptive}

The oracle acquisition rule depends on the unknown data-generating
distribution through \(J(y)\), \(I(a)\), and \(H_R\). We now show that this
dependence can be removed without sacrificing the first-order
classification-risk optimum.

Consider a two-stage procedure. In the first stage, \(m_n\) observations are
fully labeled and used to construct a pilot estimator
\(\widetilde\theta_{m_n}\), where
\begin{equation}
m_n\longrightarrow\infty,
\qquad
\frac{m_n}{n}\longrightarrow0.
\label{eq:pilot-rate}
\end{equation}
The first condition permits consistent estimation of the quantities entering
the acquisition criterion, whereas the second ensures that the pilot stage
has asymptotically negligible cost relative to the full sample.

Let \(\widehat P_m\) denote the fitted feature distribution obtained from the
pilot fit, and let
\[
\widehat J_m(y),\qquad
\widehat H_m,\qquad
\widehat I_{Y,m}
\]
denote corresponding estimators of
\[
J(y),\qquad
H_R,\qquad
I_Y(\theta_0),
\]
respectively. For an acquisition design \(a\), define
\begin{align}
\widehat I_m(a)
&=
\widehat I_{Y,m}
+
\int a(y)\widehat J_m(y)\,d\widehat P_m(y),
\label{eq:estimated-information}
\\
\widehat\Phi_m(a)
&=
\operatorname{tr}
\left\{
\widehat H_m
\widehat I_m(a)^{-1}
\right\},
\label{eq:estimated-risk-criterion}
\\
\widehat b_m(a)
&=
\int a(y)\,d\widehat P_m(y).
\label{eq:estimated-budget}
\end{align}

Because the feature distribution entering the population budget is itself
unknown, imposing the fitted constraint
\(\widehat b_m(a)\leq\rho\) need not guarantee
\(E_{\theta_0}\{a(Y)\}\leq\rho\) exactly in finite samples. We therefore use
a vanishing conservative slack. Let
\[
\varepsilon_m\downarrow0,
\]
and define the estimated design
\begin{equation}
\widehat a_m
\in
\arg\min_{0\leq a\leq1}
\widehat\Phi_m(a)
\qquad
\text{subject to}
\qquad
\widehat b_m(a)\leq\rho-\varepsilon_m.
\label{eq:adaptive-design}
\end{equation}
An \(o_p(1)\)-approximate minimizer is sufficient for all results below.

Let
\begin{equation}
\Phi_\rho^\star
=
\inf_{0\leq a\leq1,\;E_{\theta_0}\{a(Y)\}\leq\rho}
\Phi(a)
\label{eq:oracle-value}
\end{equation}
denote the oracle value. The following theorem establishes consistency of
the adaptive design at the level that is relevant for classification risk.

\begin{theorem}
\label{thm:adaptive-oracle}
Suppose the assumptions of Theorem~\ref{thm:oracle-design} hold.
Assume further that the population and fitted information matrices are
uniformly nonsingular over all measurable designs \(0\leq a\leq1\), with
probability tending to one for the fitted matrices. Assume in addition
that, as \(m\to\infty\),
\begin{align}
\sup_{0\leq a\leq1}
\left|
\widehat\Phi_m(a)-\Phi(a)
\right|
&=
o_p(\varepsilon_m),
\label{eq:uniform-criterion-consistency}
\\
\sup_{0\leq a\leq1}
\left|
\widehat b_m(a)
-
E_{\theta_0}\{a(Y)\}
\right|
&=
o_p(\varepsilon_m),
\label{eq:uniform-budget-consistency}
\end{align}
for some deterministic sequence
\(\varepsilon_m\downarrow0\).
Then the adaptive design in \eqref{eq:adaptive-design} satisfies

\begin{equation}
\Pr\left[
E_{\theta_0}
\{\widehat a_m(Y)\}
\leq\rho
\right]
\longrightarrow1,
\label{eq:adaptive-feasibility}
\end{equation}
and
\begin{equation}
\Phi(\widehat a_m)
\overset{p}{\longrightarrow}
\Phi_\rho^\star.
\label{eq:adaptive-value-consistency}
\end{equation}

Neither uniqueness of the oracle acquisition design nor convergence of
\(\widehat a_m\) to a particular oracle design is required.
\end{theorem}

The proof is given in Supplementary Section~\ref{sec:supp-adaptive}.

The distinction in the final sentence of
Theorem~\ref{thm:adaptive-oracle} is important. The statistical target is the
minimum classification-risk coefficient
\(\Phi_\rho^\star\), not a particular representation of the acquisition
function. Different acquisition designs can generate the same
classification-relevant information and hence attain the same oracle value.

The uniform conditions
\eqref{eq:uniform-criterion-consistency}--\eqref{eq:uniform-budget-consistency}
are deliberately stated at the level required by the design optimization.
Because the supremum is taken over all measurable
\(a:\mathcal Y\to[0,1]\), condition
\eqref{eq:uniform-budget-consistency} is stronger than ordinary weak
consistency of the fitted feature distribution. It is satisfied, for example,
when the fitted parametric feature distribution converges to the true feature
distribution in total variation at rate \(o_p(\varepsilon_m)\). Likewise,
\eqref{eq:uniform-criterion-consistency} follows from corresponding uniform
control of the fitted information functional, together with convergence of
\(\widehat H_m\) to \(H_R\) at the required rate and uniform nonsingularity
of the fitted information matrices. For models with smooth, regular active
Bayes faces, \(\widehat H_m\) may be obtained by evaluating the boundary
integral in \eqref{eq:risk-curvature} at the pilot estimate, provided the
estimated active faces are stable under perturbation of the fitted parameter. These conditions are sufficient rather than minimal. When the admissible
acquisition rules are restricted to a finite candidate pool or to a suitably
regular finite-dimensional class, weaker uniform empirical-process conditions
may suffice. We retain the stronger formulation here in order to state the
oracle result for the unrestricted population design problem.

We next connect design consistency to the classifier fitted after the second
stage. Conditional on the pilot data, let
\[
N_n=n-m_n
\]
denote the number of remaining observations. These observations form a fresh
sample, independent of the pilot stage, and their labels are acquired
independently according to
\[
\Delta_j\mid Y_j,\mathcal F_{m_n}
\sim
\operatorname{Bernoulli}
\{\widehat a_{m_n}(Y_j)\},
\]
where \(\mathcal F_{m_n}\) denotes the sigma-field generated by the pilot
experiment. Let \(\widehat\theta_n^{(2)}\) denote an efficient estimator based
on this second-stage experiment.

\begin{theorem}
\label{thm:adaptive-risk}
Suppose the conditions of Theorem~\ref{thm:adaptive-oracle} hold with
\(m=m_n\), where \eqref{eq:pilot-rate} is satisfied, and let
\(N_n=n-m_n\). Assume further that, conditionally on
\(\mathcal F_{m_n}\), the second-stage experiment satisfies the regular
likelihood and moment conditions required for asymptotic efficiency and for
the quadratic risk expansion, uniformly over the fitted sequence of designs.
Then
\begin{equation}
2N_n
E_{\theta_0}
\left[
R(\widehat\theta_n^{(2)})-R^*
\mid
\mathcal F_{m_n}
\right]
=
\Phi(\widehat a_{m_n})
+
o_p(1).
\label{eq:conditional-adaptive-risk}
\end{equation}
Consequently,
\begin{equation}
2N_n
E_{\theta_0}
\left[
R(\widehat\theta_n^{(2)})-R^*
\mid
\mathcal F_{m_n}
\right]
\overset{p}{\longrightarrow}
\Phi_\rho^\star.
\label{eq:oracle-risk-equivalence}
\end{equation}
Since \(N_n/n\to1\), the same first-order limit holds with \(N_n\) replaced
by \(n\).

If the sequence on the left-hand side of
\eqref{eq:conditional-adaptive-risk} is uniformly integrable, then
\begin{equation}
2N_n
\left[
E_{\theta_0}
\{R(\widehat\theta_n^{(2)})\}
-
R^*
\right]
\longrightarrow
\Phi_\rho^\star.
\label{eq:unconditional-oracle-risk}
\end{equation}
\end{theorem}

The proof is given in Supplementary Section~\ref{sec:supp-adaptive}.

Theorem~\ref{thm:adaptive-risk} shows that a vanishing pilot fraction is
sufficient for learning the classification geometry while preserving the
oracle first-order risk coefficient. The second-stage formulation makes the
conditional experiment explicit. Under standard asymptotic-linearity
conditions, the fully labeled pilot observations may also be included in
the final likelihood: because \(m_n/n\to0\), their contribution does not
alter the first-order limit.

If \(\rho n\) represents a total labeling budget that includes the pilot
sample, the second-stage acquisition fraction may be taken as
\[
\rho_n^{(2)}
=
\frac{\rho n-m_n}{n-m_n},
\]
whenever \(\rho n>m_n\). Since \(m_n/n\to0\),
\(\rho_n^{(2)}\to\rho\), so this finite-sample adjustment does not alter the
asymptotic results.

The two-stage construction is also the basic building block for a
multi-round implementation. After each batch of acquired labels, the model,
the active Bayes geometry, and the classification-value function can be
updated before the next batch is selected. The oracle result above concerns
the first-order statistical target; repeated updating is primarily a
computational device for approaching that target more accurately in finite
samples.

\section{Gaussian discriminant specialization}
\label{sec:gaussian-specialization}

We now specialize the preceding results to Gaussian discriminant analysis.
The purpose is not to restrict the general theory, but to obtain explicit
quantities that can be evaluated for linear and quadratic multiclass decision
boundaries.

Let
\[
Z\in\{1,\ldots,g\},
\qquad
\Pr(Z=k)=\pi_k,
\]
and suppose
\begin{equation}
Y\mid Z=k
\sim
N_p(\mu_k,\Sigma_k),
\qquad
k=1,\ldots,g.
\label{eq:gaussian-model}
\end{equation}
Write
\begin{equation}
r_k(y;\theta)
=
\pi_k\phi_p(y;\mu_k,\Sigma_k),
\qquad
p_\theta(y)
=
\sum_{l=1}^{g}r_l(y;\theta),
\label{eq:gaussian-weighted-density}
\end{equation}
so that
\begin{equation}
\tau_k(y;\theta)
=
\frac{r_k(y;\theta)}
{\sum_{l=1}^{g}r_l(y;\theta)}.
\label{eq:gaussian-posterior}
\end{equation}

For identifiability of the mixing proportions, use the baseline-logit
parameterization
\[
\alpha_k
=
\log\left(\frac{\pi_k}{\pi_g}\right),
\qquad
k=1,\ldots,g-1.
\]
The parameter vector is taken to be
\begin{equation}
\theta
=
\left(
\alpha^\top,
\mu_1^\top,\ldots,\mu_g^\top,
\operatorname{vech}(\Sigma_1)^\top,\ldots,
\operatorname{vech}(\Sigma_g)^\top
\right)^\top,
\label{eq:qdfa-parameter}
\end{equation}
with each \(\Sigma_k\) positive definite. The linear discriminant model is obtained by replacing the class-specific
covariance blocks by a single common covariance matrix,
\(\Sigma_1=\cdots=\Sigma_g=\Sigma\).

For each class, define the class-score vector
\begin{equation}
t_k(y;\theta)
=
\nabla_\theta
\log r_k(y;\theta),
\label{eq:class-score}
\end{equation}
and its posterior average
\begin{equation}
\bar t(y;\theta)
=
\sum_{k=1}^{g}
\tau_k(y;\theta)t_k(y;\theta).
\label{eq:posterior-class-score}
\end{equation}
Since
\[
p_\theta(y)=\sum_{k=1}^{g}r_k(y;\theta),
\]
we have
\begin{equation}
\bar t(y;\theta)
=
\nabla_\theta\log p_\theta(y).
\label{eq:feature-score-class-score}
\end{equation}
Moreover,
\begin{equation}
\nabla_\theta
\log\tau_k(y;\theta)
=
t_k(y;\theta)-\bar t(y;\theta).
\label{eq:posterior-score}
\end{equation}

It follows immediately that the conditional information supplied by the
class label has the form
\begin{equation}
J(y)
=
\sum_{k=1}^{g}
\tau_k(y)
\left\{
t_k(y)-\bar t(y)
\right\}
\left\{
t_k(y)-\bar t(y)
\right\}^{\top}.
\label{eq:gaussian-J}
\end{equation}
Thus \(J(y)\) is the posterior covariance matrix of the class-score vector
\(t_Z(y)\).

The corresponding feature information is
\begin{equation}
I_Y
=
E
\left[
\bar t(Y)\bar t(Y)^\top
\right],
\label{eq:gaussian-feature-information}
\end{equation}
and the information under an acquisition design \(a\) becomes
\begin{equation}
I(a)
=
I_Y
+
E
\left[
a(Y)
\sum_{k=1}^{g}
\tau_k(Y)
\{t_k(Y)-\bar t(Y)\}
\{t_k(Y)-\bar t(Y)\}^{\top}
\right].
\label{eq:gaussian-design-information}
\end{equation}

The class-score vectors in \eqref{eq:class-score} are available explicitly.
Let
\[
e_k^{(\pi)}
=
\nabla_\alpha\log\pi_k.
\]
Then
\begin{equation}
e_k^{(\pi)}
=
\begin{cases}
e_k-\pi_{1:g-1}, & k<g,\\[3pt]
-\pi_{1:g-1}, & k=g,
\end{cases}
\label{eq:prior-score}
\end{equation}
where \(e_k\) is the \(k\)th coordinate vector in
\(\mathbb R^{g-1}\).
For the Gaussian parameters,
\begin{equation}
\nabla_{\mu_k}
\log r_k(y;\theta)
=
\Sigma_k^{-1}(y-\mu_k),
\label{eq:mean-score}
\end{equation}
and
\begin{equation}
\nabla_{\operatorname{vech}(\Sigma_k)}
\log r_k(y;\theta)
=
\frac12
D_p^\top
\operatorname{vec}
\left[
\Sigma_k^{-1}
\left\{
(y-\mu_k)(y-\mu_k)^\top-\Sigma_k
\right\}
\Sigma_k^{-1}
\right],
\label{eq:covariance-score}
\end{equation}
where \(D_p\) is the duplication matrix. Derivatives with respect to the
parameters of classes other than \(k\) are zero. These expressions provide
the complete vector \(t_k(y;\theta)\).

We next turn to the classification-risk curvature. For classes \(k\) and
\(l\), define
\[
g_{kl}(y;\theta)
=
r_k(y;\theta)-r_l(y;\theta).
\]
On the pairwise Bayes boundary
\[
g_{kl}(y;\theta_0)=0,
\]
we have
\[
r_k(y;\theta_0)
=
r_l(y;\theta_0).
\]
Write their common value as
\[
r_{kl}(y).
\]
Using
\[
\nabla_\theta r_k(y;\theta)
=
r_k(y;\theta)t_k(y;\theta),
\]
the parameter derivative of the pairwise density contrast simplifies on the
boundary to
\begin{equation}
b_{kl}(y)
=
r_{kl}(y)
\left\{
t_k(y)-t_l(y)
\right\}.
\label{eq:gaussian-boundary-parameter-gradient}
\end{equation}

For the feature derivative,
\[
\nabla_y\log r_k(y)
=
-\Sigma_k^{-1}(y-\mu_k),
\]
so that, on the same boundary,
\begin{equation}
\nabla_y g_{kl}(y;\theta_0)
=
r_{kl}(y)d_{kl}(y),
\label{eq:gaussian-boundary-feature-gradient}
\end{equation}
where
\begin{equation}
d_{kl}(y)
=
\Sigma_l^{-1}(y-\mu_l)
-
\Sigma_k^{-1}(y-\mu_k).
\label{eq:normal-direction}
\end{equation}

Substituting
\eqref{eq:gaussian-boundary-parameter-gradient} and
\eqref{eq:gaussian-boundary-feature-gradient} into the general curvature
formula gives the following useful representation.

\begin{proposition}
\label{prop:gaussian-face-curvature}
For the Gaussian discriminant model
\eqref{eq:gaussian-model}, suppose the active pairwise face
\(\mathcal F_{kl}\) is regular, so that
\(d_{kl}(s)\neq0\) on \(\mathcal F_{kl}\). Then
\begin{equation}
H_{kl}
=
\int_{\mathcal F_{kl}}
r_{kl}(s)
\frac{
\delta t_{kl}(s)
\delta t_{kl}(s)^\top
}{
\|d_{kl}(s)\|
}
\,dS(s),
\label{eq:gaussian-face-curvature}
\end{equation}
where
\begin{equation}
\delta t_{kl}(s)
=
t_k(s)-t_l(s).
\label{eq:class-score-contrast}
\end{equation}
Consequently,
\[
H_R
=
\sum_{k<l}H_{kl},
\]
where the sum is taken over the active pairwise Bayes faces.
\end{proposition}

The derivation is given in Supplementary Section~\ref{sec:supp-gaussian}.

Equation~\eqref{eq:gaussian-face-curvature} is especially useful
computationally. The factor \(r_{kl}(s)\) weights locations according to the
probability density present on the decision boundary, while
\(\delta t_{kl}(s)\) describes the parameter directions in which the
\(k\)--\(l\) contrast changes. The denominator
\(\|d_{kl}(s)\|\) accounts for the local rate at which the two weighted class
densities separate in the direction normal to the boundary.

The acquisition score also admits a convenient scalar representation. Define
\begin{equation}
G_a
=
I(a)^{-1}H_RI(a)^{-1}
\label{eq:gaussian-G}
\end{equation}
and
\[
u_k(y)
=
t_k(y)-\bar t(y).
\]
Then, using \eqref{eq:gaussian-J},
\begin{align}
\psi_a(y)
&=
\operatorname{tr}\{G_aJ(y)\}
\nonumber\\
&=
\sum_{k=1}^{g}
\tau_k(y)
u_k(y)^\top
G_a
u_k(y).
\label{eq:gaussian-value-score}
\end{align}
Thus the classification value at \(y\) can be evaluated without explicitly
forming a large matrix product for every candidate observation.

Likewise, with
\[
G_{kl,a}
=
I(a)^{-1}H_{kl}I(a)^{-1},
\]
the face-specific value is
\begin{equation}
\psi_{kl,a}(y)
=
\sum_{r=1}^{g}
\tau_r(y)
u_r(y)^\top
G_{kl,a}
u_r(y).
\label{eq:gaussian-face-value-score}
\end{equation}

These formulas reveal a useful distinction between posterior ambiguity and
classification value. The posterior probabilities \(\tau_k(y)\) determine
how uncertain the class membership is, but the quadratic forms
\[
u_k(y)^\top G_a u_k(y)
\]
determine whether the information associated with each possible class label
is directed toward parameter combinations that influence the active
multiclass decision boundary. Two observations having nearly identical
posterior probability vectors can therefore receive substantially different
classification values.

For linear discriminant analysis, the pairwise Bayes faces are hyperplanes
and the integrations in \eqref{eq:gaussian-face-curvature} simplify
accordingly. Under quadratic discriminant analysis, unequal covariance
matrices generate curved pairwise faces, so the full geometric structure of
\(H_R\) is retained. For this reason, quadratic discrimination provides the
principal numerical setting considered below.

\section{Numerical study}
\label{sec:numerical}

We examine the finite-sample behavior of classification-risk-optimal
label acquisition in two three-class Gaussian discriminant settings.
The first setting has strongly heterogeneous covariance matrices and
curved Bayes boundaries, so that the distinction between posterior
uncertainty and classification value is pronounced. The second retains
the same class proportions and means but substantially reduces the
covariance heterogeneity, producing a near-LDA geometry in which the
active boundaries are much closer to linear. The second setting therefore
serves as a control regime for assessing whether the behavior of the
proposed acquisition rule depends on the classification geometry.

\subsection{Simulation design}
\label{subsec:simulation-design}

In both scenarios there are \(g=3\) classes and \(p=2\) observed
features, with class probabilities
\[
\boldsymbol{\pi}
=
(0.36,0.34,0.30)
\]
and mean vectors
\[
\boldsymbol{\mu}_1=(-1.60,0)^\top,
\qquad
\boldsymbol{\mu}_2=(1.50,0.20)^\top,
\qquad
\boldsymbol{\mu}_3=(0,1.90)^\top.
\]

For each Monte Carlo replication, a training sample of size \(n=1000\)
was generated. An initial random pilot sample of 60 observations was
labeled and used to fit the QDA model. Additional labels were then
selected so that the total labeled proportions were
\[
\rho\in\{0.10,0.20,0.30\}.
\]
Both the pilot and post-acquisition QDA fits were obtained by
semi-supervised maximum likelihood using the EM algorithm. The observed
class memberships were held fixed, whereas the memberships of the
unlabeled observations were treated as latent and updated through their
posterior class probabilities in the E-step; the class proportions, means,
and class-specific covariance matrices were updated in the M-step.
After acquisition, the QDA model was refitted using all observed feature
vectors together with the labels available under the corresponding
design. Classification performance was evaluated on a common test sample
of size \(100000\). Each experiment was repeated 100 times.

We compare six acquisition strategies. Random acquisition samples the
additional labels uniformly from the candidate pool. Entropy and margin
sampling use the fitted posterior class probabilities. The Fisher design
targets global parameter information through
\[
\Phi_{\mathrm F}(a)
=
\operatorname{tr}
\left\{
I_{\mathrm{CC}}I(a)^{-1}
\right\},
\]
where \(I_{\mathrm{CC}}\) denotes the complete-classification Fisher
information. The proposed adaptive risk-optimal rule instead targets
\[
\Phi_R(a)
=
\operatorname{tr}
\left\{
H_R I(a)^{-1}
\right\},
\]
with the model-dependent quantities evaluated at the pilot estimate.
Because all feature vectors in the candidate pool are observed before label
acquisition, population expectations entering the design criterion are
replaced in the finite-pool implementation by empirical averages over the
observed feature pool. For reference, we also include an oracle risk-optimal
design in which the model-dependent quantities are evaluated at the true
generating parameter.
The oracle rule is not implementable and is included only to assess the
finite-sample effect of estimating the classification-risk geometry.

The term ``oracle'' refers to optimization of the population leading-risk
criterion at the true generating parameter. It does not imply the smallest
realized finite-sample classification error in every Monte Carlo setting.
Consequently, small finite-sample reversals between the adaptive and oracle
procedures are compatible with the theoretical oracle property.

The finite candidate-pool optimization was performed over the relaxed
weights \(0\leq a_i\leq1\), followed by selection of the observations
with the largest weights. This approximation was numerically negligible.
In the population illustration for Scenario~1, only 6 of 5000 weights
were fractional and the corresponding rounding loss in the objective was
\(0.0002\%\). Across both simulation scenarios, fractional weights formed
only a small proportion of the candidate pool and the induced objective
loss remained negligible.

\subsection{Scenario 1: heterogeneous QDA geometry}
\label{subsec:scenario1}

The covariance matrices in the first scenario were
\[
\boldsymbol{\Sigma}_1=
\begin{pmatrix}
1.00 & 0.45\\
0.45 & 0.70
\end{pmatrix},
\qquad
\boldsymbol{\Sigma}_2=
\begin{pmatrix}
0.80 & -0.35\\
-0.35 & 1.10
\end{pmatrix},
\qquad
\boldsymbol{\Sigma}_3=
\begin{pmatrix}
1.25 & 0.15\\
0.15 & 0.55
\end{pmatrix}.
\]
These matrices generate clearly curved active Bayes faces and substantial
heterogeneity in the directions through which the class labels contribute
information.

\begin{figure}[H]
    \centering
    \includegraphics[width=\textwidth]{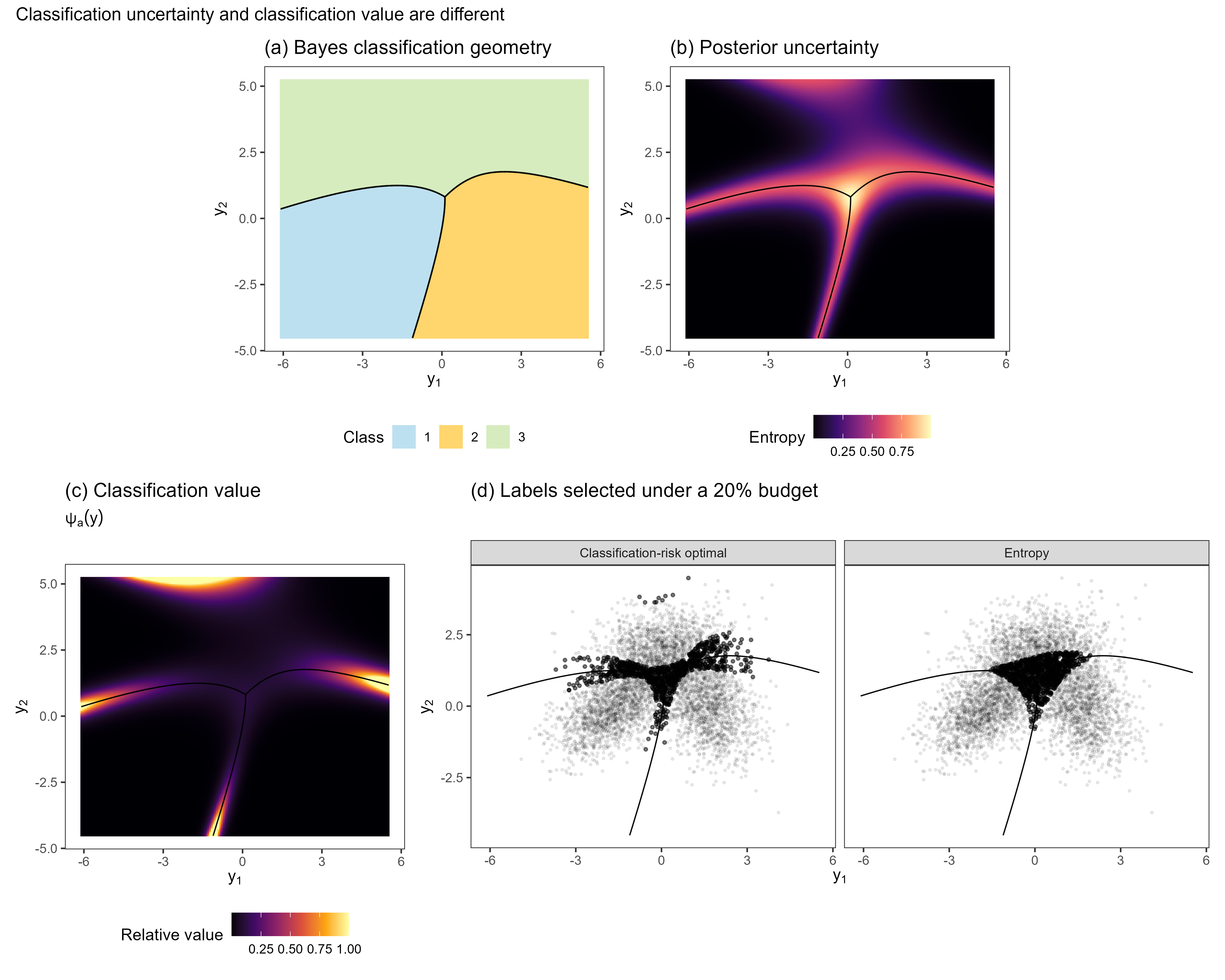}
    \caption{
    Classification geometry in Scenario~1.
    Panel (a) shows the Bayes classification regions and active pairwise
    boundaries. Panel (b) shows normalized posterior entropy and panel
    (c) the classification-value function \(\psi_a(y)\) under a uniform
    \(20\%\) acquisition design. Panel (d) compares the locations selected
    by classification-risk-optimal and entropy acquisition under the same
    labeling budget; gray points show the candidate feature population.
    }
    \label{fig:scenario1-geometry}
\end{figure}

Figure~\ref{fig:scenario1-geometry} illustrates the resulting population
geometry. Posterior entropy is concentrated around regions in which the
largest class probabilities are similar. The classification-value
surface \(\psi_a(y)\) has a visibly different structure because it also
incorporates the orientation of the conditional label information
relative to parameter directions that perturb the active Bayes faces.
The distinction is especially clear in the locations selected under a
\(20\%\) budget. Entropy sampling concentrates observations around the
most ambiguous portions of the boundaries, whereas classification-risk
acquisition extends along the active faces and selects observations that
are informative for their movement and deformation.

The estimated Bayes error in this scenario was \(0.14845\).
The first block of Table~\ref{tab:simulation-results} summarizes the
classification results. At a \(10\%\) labeling budget, the adaptive
risk-optimal rule attained a mean error of \(0.15138\), compared with
\(0.15175\) for entropy, \(0.15159\) for margin, \(0.15156\) for the
Fisher design, and \(0.15209\) for random acquisition. Its performance
was close to the oracle benchmark of \(0.15128\).

At the \(20\%\) budget, the adaptive design again had the lowest mean
error among the implementable strategies, \(0.15020\), compared with
\(0.15031\) for Fisher, \(0.15037\) for margin, \(0.15054\) for entropy,
and \(0.15109\) for random acquisition. When \(30\%\) of the labels were
available, the differences among the targeted acquisition methods became
small, with mean errors between \(0.14980\) and \(0.14992\), while random
acquisition remained less accurate at \(0.15043\).

\subsection{Scenario 2: weak covariance heterogeneity}
\label{subsec:scenario2}

The second scenario keeps the class probabilities and means unchanged
but replaces the covariance matrices by
\[
\boldsymbol{\Sigma}_1=
\begin{pmatrix}
1.00 & 0.15\\
0.15 & 0.90
\end{pmatrix},
\qquad
\boldsymbol{\Sigma}_2=
\begin{pmatrix}
0.95 & 0.12\\
0.12 & 0.95
\end{pmatrix},
\qquad
\boldsymbol{\Sigma}_3=
\begin{pmatrix}
1.05 & 0.18\\
0.18 & 0.85
\end{pmatrix}.
\]
The covariance matrices are now close to one another, so the resulting
Bayes boundaries are substantially closer to linear than in
Scenario~1.

\begin{figure}[H]
    \centering
    \includegraphics[width=\textwidth]{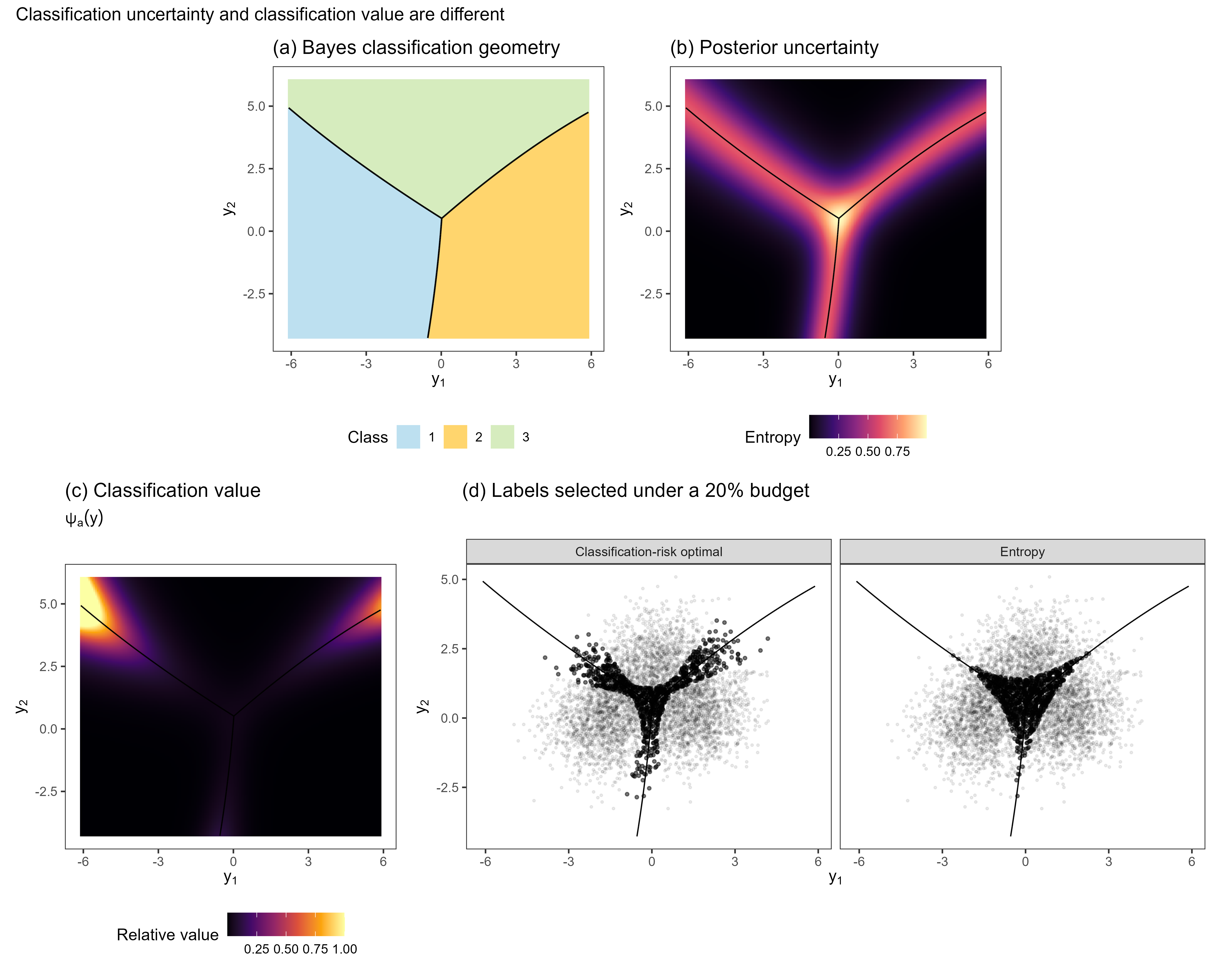}
    \caption{
    Classification geometry in Scenario~2 with weak covariance
    heterogeneity. The panels have the same interpretation as in
    Figure~\ref{fig:scenario1-geometry}. The active Bayes boundaries are
    now substantially closer to linear, providing a control regime in
    which posterior uncertainty and classification value are more closely
    aligned.
    }
    \label{fig:scenario2-geometry}
\end{figure}

Figure~\ref{fig:scenario2-geometry} confirms this change in geometry.
The three pairwise boundaries form an almost linear three-way partition.
Posterior entropy and classification value are consequently more closely
aligned than in Scenario~1, although they are still not identical.
Classification-risk acquisition spreads selected observations farther
along the active faces, whereas entropy sampling remains more concentrated
around regions of highest posterior ambiguity.

The estimated Bayes error for Scenario~2 was \(0.15882\).
The second block of Table~\ref{tab:simulation-results} shows that the
adaptive risk-optimal rule remained competitive under this weaker
geometric heterogeneity. At a \(10\%\) budget, its mean classification
error was \(0.16193\), compared with \(0.16290\) for entropy,
\(0.16253\) for margin, \(0.16205\) for Fisher, and \(0.16316\) for
random acquisition. At the \(20\%\) budget, the corresponding error was
\(0.16046\), slightly below Fisher at \(0.16049\), margin at \(0.16080\),
entropy at \(0.16142\), and random acquisition at \(0.16177\).

With \(30\%\) of the labels observed, the targeted methods again became
more similar. The adaptive and oracle risk-optimal rules both had mean
error approximately \(0.16002\), while Fisher, margin, and entropy gave
\(0.16014\), \(0.16023\), and \(0.16043\), respectively. Random
acquisition remained less accurate at \(0.16111\).

\begin{table}[t]
\centering
\caption{
Mean test classification error over 100 Monte Carlo replications.
Monte Carlo standard errors are shown in parentheses. The smallest mean
error among the implementable acquisition rules within each labeling
budget is shown in bold; the oracle risk-optimal rule is included only as
a theoretical benchmark.
}
\label{tab:simulation-results}
\begin{tabular}{lccc}
\hline
Acquisition rule
& \(10\%\)
& \(20\%\)
& \(30\%\) \\
\hline
\multicolumn{4}{l}{\textit{Scenario 1: heterogeneous QDA}}\\[1mm]

Random
& 0.15209 (0.000258)
& 0.15109 (0.000175)
& 0.15043 (0.000128) \\

Entropy
& 0.15175 (0.000250)
& 0.15054 (0.000172)
& 0.14992 (0.000087) \\

Margin
& 0.15159 (0.000183)
& 0.15037 (0.000120)
& 0.14987 (0.000084) \\

Fisher
& 0.15156 (0.000270)
& 0.15031 (0.000125)
& 0.14989 (0.000085) \\

Adaptive risk-optimal
& \textbf{0.15138} (0.000232)
& \textbf{0.15020} (0.000103)
& \textbf{0.14980} (0.000082) \\

Oracle risk-optimal
& 0.15128 (0.000199)
& 0.15026 (0.000113)
& 0.14983 (0.000079) \\

\hline
\multicolumn{4}{l}{\textit{Scenario 2: weak covariance heterogeneity}}\\[1mm]

Random
& 0.16316 (0.000358)
& 0.16177 (0.000203)
& 0.16111 (0.000139) \\

Entropy
& 0.16290 (0.000303)
& 0.16142 (0.000197)
& 0.16043 (0.000109) \\

Margin
& 0.16253 (0.000275)
& 0.16080 (0.000140)
& 0.16023 (0.000099) \\

Fisher
& 0.16205 (0.000256)
& 0.16049 (0.000121)
& 0.16014 (0.000115) \\

Adaptive risk-optimal
& \textbf{0.16193} (0.000232)
& \textbf{0.16046} (0.000116)
& \textbf{0.16002} (0.000083) \\

Oracle risk-optimal
& 0.16199 (0.000239)
& 0.16060 (0.000117)
& 0.16002 (0.000089) \\

\hline
\end{tabular}
\end{table}

\begin{figure}[H]
    \centering
    \includegraphics[width=\textwidth]{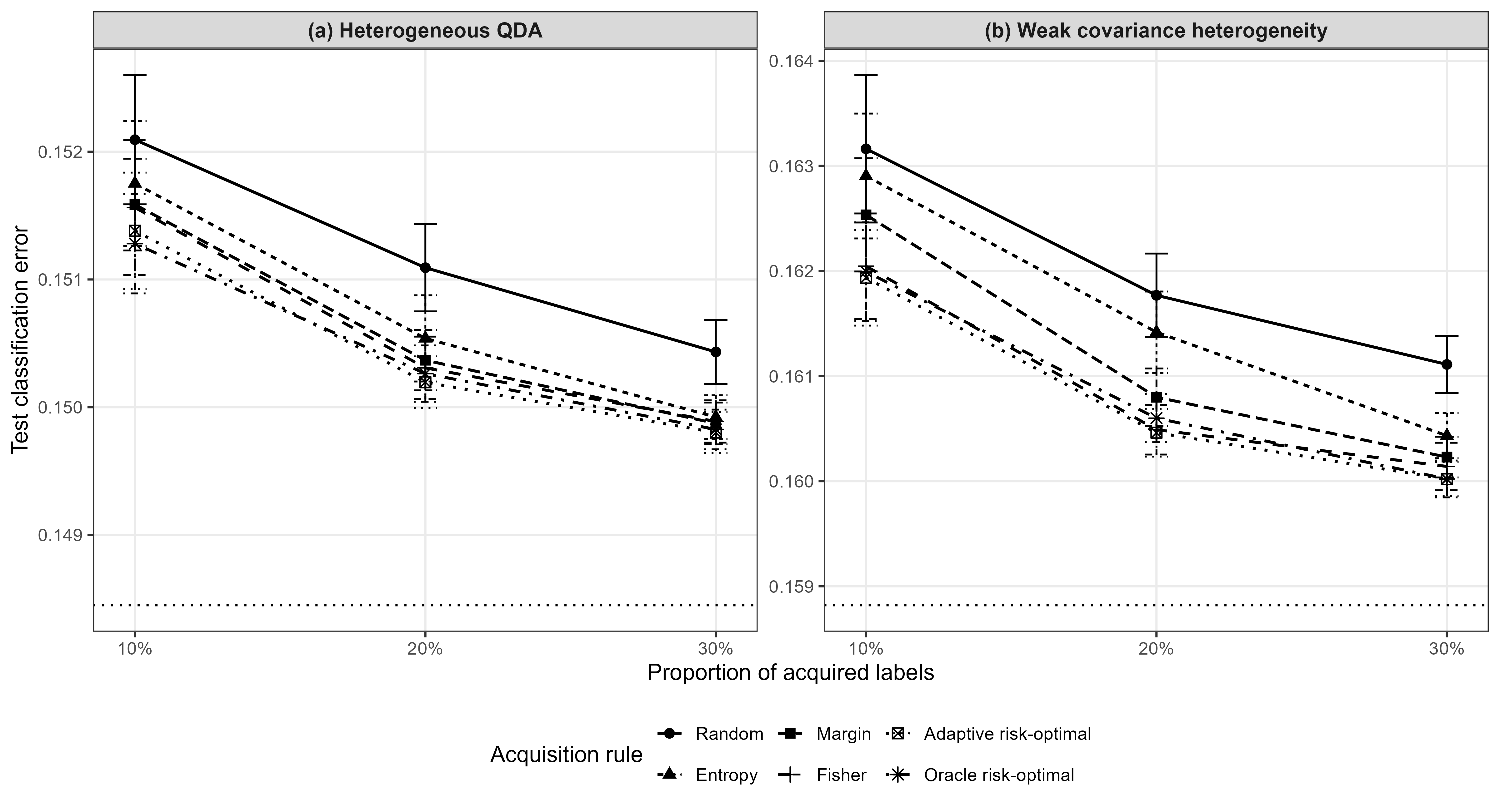}
    \caption{
    Mean test classification error as a function of the total proportion
    of acquired labels for (a) Scenario~1 with heterogeneous QDA geometry
    and (b) Scenario~2 with weak covariance heterogeneity. Error bars are
    \(95\%\) Monte Carlo intervals based on 100 replications. The
    horizontal dotted lines indicate the corresponding estimated Bayes
    errors, \(0.14845\) in Scenario~1 and \(0.15882\) in Scenario~2.
    }
    \label{fig:simulation-performance}
\end{figure}

Figure~\ref{fig:simulation-performance} compares the Monte Carlo
classification results directly across the two scenarios. In both
settings, random acquisition has the largest mean classification error,
while the differences among the targeted acquisition rules decrease as
the labeling budget increases. In Scenario~1, the adaptive
classification-risk rule has the smallest mean error among the
implementable procedures at each budget. The same pattern is retained
in Scenario~2, although the differences between the adaptive
risk-optimal and Fisher designs are substantially smaller, consistent
with the weaker covariance heterogeneity and the nearly linear Bayes
geometry.

Overall, the two scenarios support the geometric interpretation of the
acquisition criterion rather than a uniform dominance claim. Under the
strongly heterogeneous QDA geometry of Scenario~1, posterior uncertainty
and classification value generate visibly different acquisition
patterns, and the adaptive risk-optimal rule gives the smallest mean
classification error among the implementable procedures at all three
budgets. Under the weaker covariance heterogeneity of Scenario~2, the
differences between the risk-optimal and Fisher designs become much
smaller, while both remain more effective than random acquisition and,
particularly at the smaller budgets, uncertainty-based acquisition.
These results are consistent with the role of \(H_R\): the benefit of
classification-risk targeting depends on how strongly the available
label information differs in its alignment with parameter directions
that perturb the active Bayes boundary.


\section{Real-data application}
\label{sec:realdata}

As a real-data application, we considered the Statlog Landsat Satellite
data set of \citet{Srinivasan1993}. The data consist of multispectral measurements
from Landsat imagery, with the objective of classifying the land-cover type
of the central pixel in a \(3\times3\) neighborhood. The benchmark contains
4,435 training observations and 2,000 test observations. Six land-cover
classes are represented: red soil, cotton crop, grey soil, damp grey soil,
soil with vegetation stubble, and very damp grey soil. Class 6 in the
original coding is absent from the released data.

Following the recommendation accompanying the data set, we used the four
spectral measurements corresponding to the central pixel, namely attributes
17--20. This avoids introducing neighboring-pixel measurements from
\(3\times3\) windows that may straddle a land-cover boundary. The four
predictors were standardized using the full benchmark training feature pool,
and the resulting transformation was applied to the fixed benchmark test
set. Since the training features are observed before label acquisition, this
unsupervised preprocessing does not use unavailable class-label information.

The supplied benchmark training--test split was retained throughout the
analysis. The 50 replications below do not correspond to repeated train--test
partitions.
Instead, each replication uses the same 4,435 training features and 2,000
test observations but generates a new random pilot labeling set and the
corresponding acquisition decisions. An initial random pilot comprising
5\% of the training pool was labeled, giving 222 pilot labels. A
semi-supervised six-class Gaussian QDA model was then fitted using the
observed pilot labels together with all unlabeled training features.
Estimation was carried out by maximum likelihood using the EM algorithm,
with the observed class memberships held fixed and the memberships of the
unlabeled observations treated as latent. Posterior class probabilities
for the unlabeled observations were updated in the E-step, and the class
proportions, means, and class-specific covariance matrices were updated
in the M-step. A small eigenvalue floor was applied to the covariance
matrices for numerical stability.

We considered total labeling budgets of 10\%, 20\%, and 30\%, corresponding to 444, 887, and 1,331 labeled training observations, respectively.
Including the common pilot, these budgets require the acquisition of an
additional 222, 665, and 1,109 labels. The adaptive classification-risk design was compared with random
acquisition, entropy sampling, margin sampling, and the Fisher-information
design. Each
target budget was constructed independently from the same pilot fit within
a replication.

For the Fisher and adaptive designs, the relaxed finite-pool criterion was
solved by the Frank--Wolfe algorithm and converted to an exact labeling set
by deterministic top-budget rounding. All 300 fitted design problems
converged to the prescribed relative Frank--Wolfe tolerance of \(10^{-5}\),
and no model-fitting failures occurred. The rounding approximation was also
negligible: the largest observed increase in the relaxed objective was about
0.021\%, occurring for the adaptive design at the 10\% budget.

\begin{figure}[H]
\centering
\includegraphics[width=\textwidth]{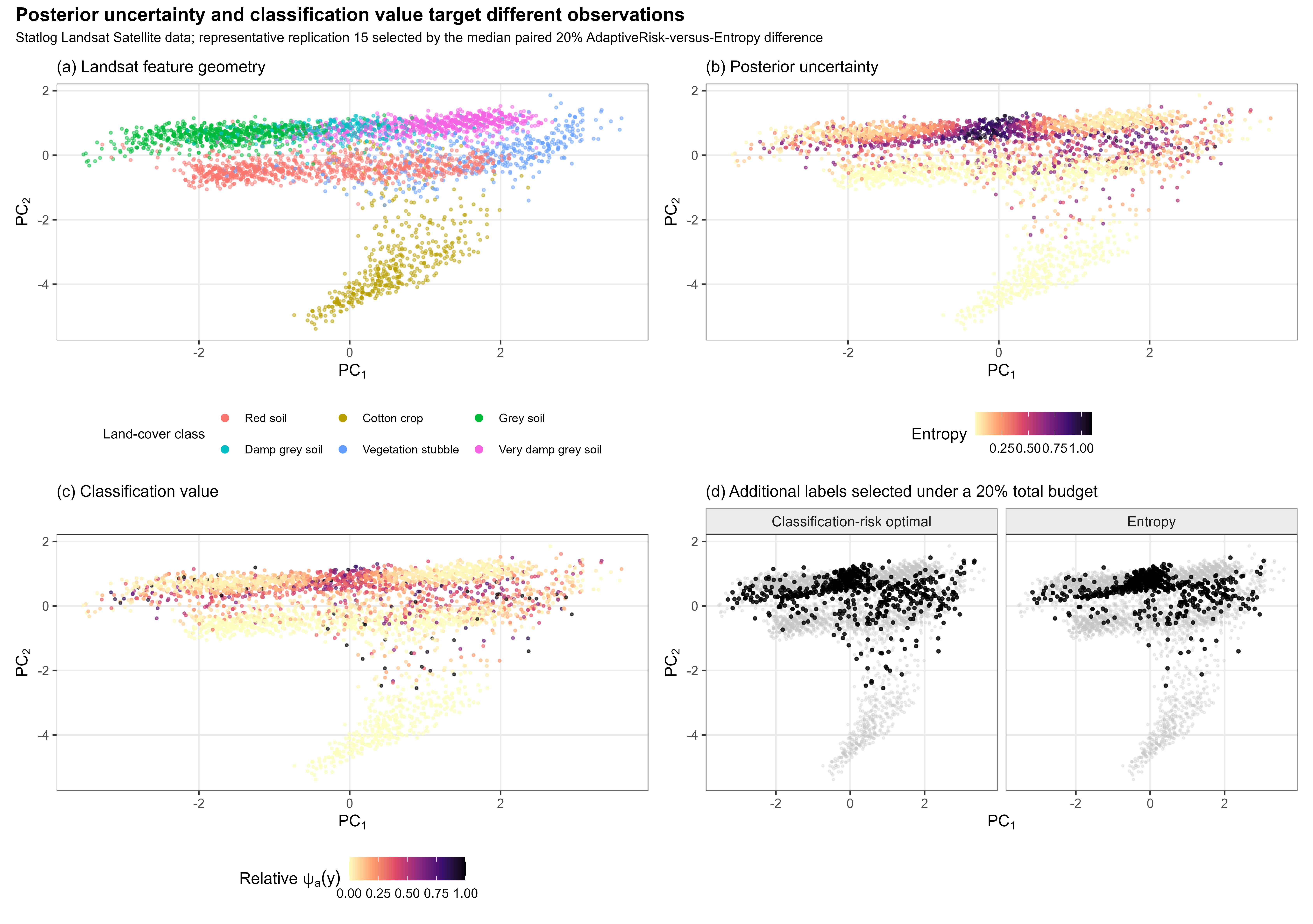}
\caption{
Acquisition geometry for the Statlog Landsat Satellite data in representative
replication 15. Panel (a) displays the six land-cover classes in the first
two principal components of the four standardized central-pixel spectral
measurements. Panel (b) shows posterior classification uncertainty measured
by entropy, whereas panel (c) shows the estimated relative classification
value \(\widehat{\psi}_a(y)\). Panel (d) compares the additional observations
selected by classification-risk-optimal and entropy acquisition under a
20\% total labeling budget; black points denote acquired labels and gray
points the remaining training pool. The principal components are used only
for visualization: posterior probabilities, classification values, and
acquisition decisions are computed from the full four-dimensional QDA model.
The replication was selected as the one whose paired
AdaptiveRisk-versus-Entropy difference was closest to the median across the
50 replications.
}
\label{fig:landsat-geometry}
\end{figure}

Figure~\ref{fig:landsat-geometry} illustrates the acquisition geometry using
an objectively selected replication at the 20\% labeling budget. For each
of the 50 replications, we computed the paired difference between entropy
and adaptive classification-risk test error and selected the replication
whose difference was closest to the median. This criterion selected
replication 15, for which the adaptive and entropy test errors were 0.1550
and 0.1555, respectively. The first two principal components of the four
standardized spectral variables are used only to display the observations;
the fitted QDA model, posterior probabilities, classification values, and
acquisition decisions are all computed from the full four-dimensional
spectral representation.

The distinction between posterior uncertainty and classification value is
particularly visible in Figure~\ref{fig:landsat-geometry}. Posterior entropy
is concentrated mainly in regions where several fitted land-cover classes
overlap, especially along the dense upper portion of the projected feature
space. The classification-value surface is more heterogeneous within these
same regions and also assigns appreciable value to observations that are not
among the most uncertain according to posterior entropy. Consequently, the
two rules do not simply recover the same ranking of the unlabeled pool.
Figure~\ref{fig:landsat-geometry}(d) shows that entropy acquisition remains
strongly concentrated in the principal uncertainty regions, whereas the
classification-risk rule distributes part of its labeling budget differently
across the feature geometry. This provides a real-data illustration that predictive uncertainty and
value for reducing classification risk are distinct quantities.

Table~\ref{tab:landsat-results} reports the mean benchmark test errors across
the 50 pilot/acquisition replications. Monte Carlo standard errors quantify
variation induced by the random pilot and acquisition experiment conditional
on the fixed benchmark training and test samples.

\begin{table}[t]
\centering
\caption{
Mean benchmark test classification error and balanced error for the Statlog
Landsat Satellite data over 50 repeated pilot/acquisition replications.
Monte Carlo standard errors are reported in parentheses. The smallest mean
error within each labeling budget is shown in bold.
}
\label{tab:landsat-results}

\resizebox{\textwidth}{!}{%
\begin{tabular}{lcccccc}
\hline
& \multicolumn{2}{c}{10\%}
& \multicolumn{2}{c}{20\%}
& \multicolumn{2}{c}{30\%} \\
\cline{2-3}\cline{4-5}\cline{6-7}
Method
& Error & Balanced
& Error & Balanced
& Error & Balanced \\
\hline
Random
& 0.18511 (0.00296)
& 0.22786 (0.00495)
& 0.16159 (0.00129)
& 0.19480 (0.00218)
& 0.15639 (0.00090)
& \textbf{0.19245} (0.00144)
\\
Entropy
& \textbf{0.17098} (0.00264)
& \textbf{0.20396} (0.00328)
& 0.15709 (0.00055)
& 0.19310 (0.00085)
& 0.15682 (0.00026)
& 0.19549 (0.00042)
\\
Margin
& 0.17250 (0.00248)
& 0.21120 (0.00316)
& \textbf{0.15626} (0.00050)
& \textbf{0.19256} (0.00084)
& \textbf{0.15638} (0.00024)
& 0.19483 (0.00044)
\\
Fisher
& 0.17822 (0.00255)
& 0.22160 (0.00444)
& 0.15835 (0.00041)
& 0.19474 (0.00076)
& 0.15716 (0.00029)
& 0.19573 (0.00052)
\\
Adaptive risk
& 0.17470 (0.00255)
& 0.21620 (0.00438)
& 0.15675 (0.00042)
& 0.19271 (0.00072)
& 0.15657 (0.00030)
& 0.19511 (0.00048)
\\
\hline
\end{tabular}%
}
\end{table}

\begin{figure}[H]
\centering
\includegraphics[width=0.82\textwidth]{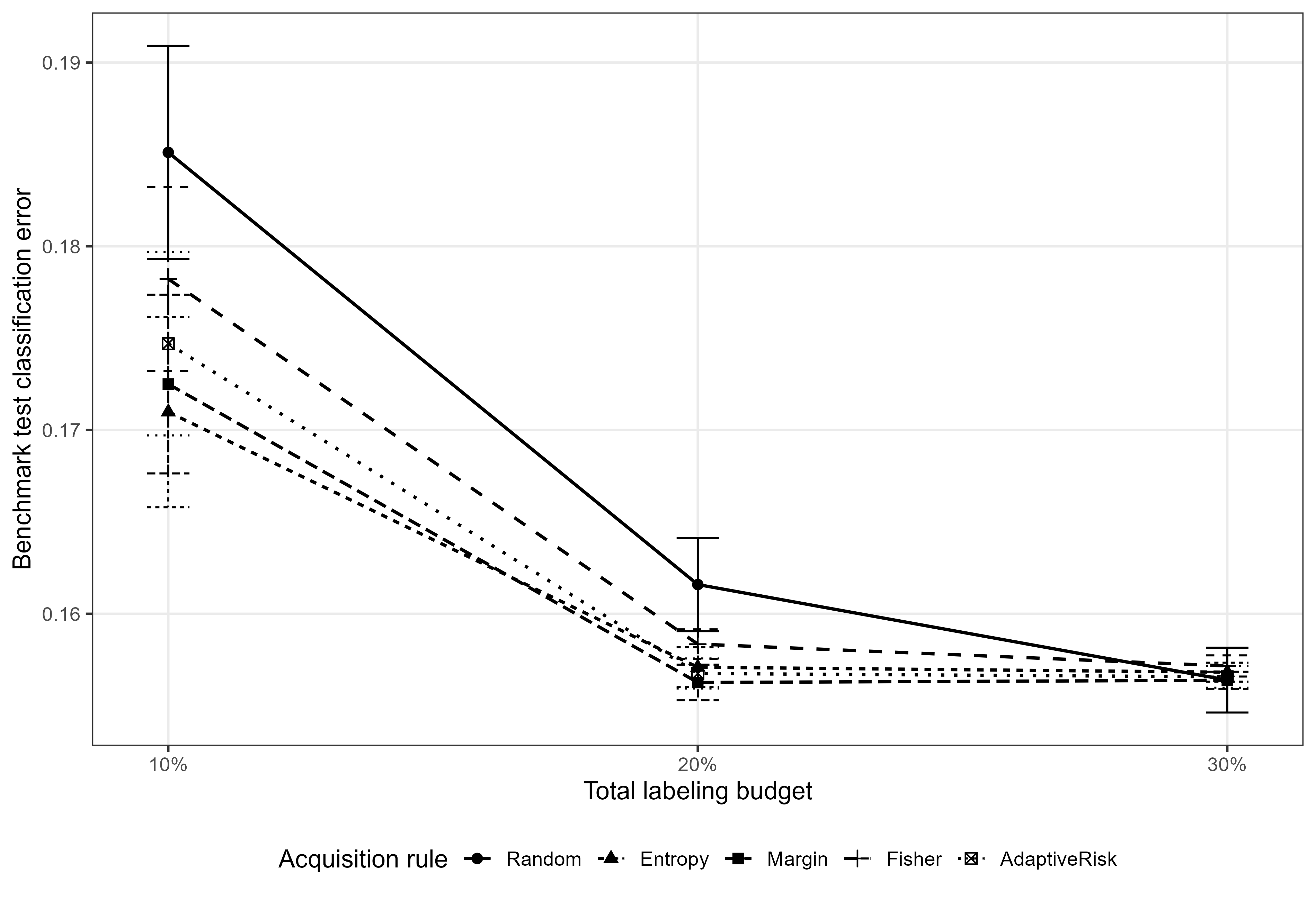}
\caption{
Mean benchmark test classification error for the Statlog Landsat Satellite
data over 50 repeated pilot/acquisition replications. Error bars represent
\(\pm1.96\) Monte Carlo standard errors. The benchmark training and test
sets are fixed across replications; the random pilot labels and resulting
acquisition decisions vary between replications.
}
\label{fig:landsat-error}
\end{figure}

At the 10\% total labeling budget, all targeted acquisition rules improved
substantially on random labeling. The adaptive classification-risk rule
reduced the mean test error from 0.1851 under random acquisition to 0.1747,
an absolute reduction of 0.0104, or approximately 5.6\% relative to the
random-acquisition error. The paired 95\% Monte Carlo interval for the
random-minus-adaptive difference was \((0.0050,0.0158)\). The
Fisher-information design had a larger mean error, 0.1782, than the adaptive
rule, although their paired difference of 0.0035 was not clearly
distinguishable from zero at this budget. Entropy and margin sampling
attained smaller mean errors, 0.1710 and 0.1725, respectively. Thus the
smallest-budget experiment shows a clear advantage of targeted acquisition
over random labeling, but not uniform superiority of the classification-risk
criterion over uncertainty sampling.

At the 20\% labeling budget, the differences among the targeted rules became
substantially smaller. Margin sampling attained the smallest mean test error,
0.15626, followed closely by the adaptive classification-risk rule at
0.15675 and entropy sampling at 0.15709. The paired difference between
entropy and the adaptive rule was only 0.00034, with a 95\% Monte Carlo
interval of \((-0.00049,0.00117)\), and the corresponding comparison with
margin sampling was similarly small. The distinction from the
Fisher-information criterion was clearer: Fisher attained a mean error of
0.15835, giving a paired Fisher-minus-adaptive difference of 0.00160 with
95\% Monte Carlo interval \((0.00097,0.00223)\). The adaptive rule also
improved on random acquisition by 0.00484, with interval
\((0.00239,0.00729)\).

At the 30\% budget, the mean overall errors of all five procedures were
within approximately \(8\times10^{-4}\) of one another. Margin sampling had
the smallest mean error, 0.15638, followed almost identically by random
acquisition at 0.15639 and the adaptive rule at 0.15657. Entropy sampling
and Fisher information gave mean errors of 0.15682 and 0.15716,
respectively. Although the absolute differences were very small, the paired
Fisher-minus-adaptive difference remained positive, 0.00059, with a 95\%
Monte Carlo interval of \((0.00020,0.00098)\). This convergence of predictive
performance at the largest labeling budget is consistent with the diminishing
importance of acquisition strategy once a substantial fraction of the
training labels has been observed.

The balanced-error results reinforce the main conclusion but also reveal
some class heterogeneity. At the 10\% budget, entropy sampling attained the
smallest balanced error, whereas at 20\% the margin and adaptive rules were
nearly indistinguishable. In particular, the adaptive design improved upon
the Fisher-information design in balanced error at 20\% by approximately
0.00203, with paired 95\% Monte Carlo interval
\((0.00095,0.00312)\). At 30\%, differences among the targeted rules were
again very small, although random acquisition had the smallest mean balanced
error. The discrepancy between overall and balanced error at the largest
budget indicates that the remaining differences are partly associated with
the unequal difficulty and prevalence of the six land-cover classes.

The class-specific results provide additional insight into the
classification-risk versus Fisher-information comparison. At the 20\%
budget, the differences between the two designs were heterogeneous across
the six land-cover classes, with the adaptive rule attaining smaller mean
class-specific error for several classes and Fisher performing better for
others. This heterogeneity is consistent with the mechanism represented by
\(H_R\): classification-risk acquisition weights information according to
its relevance for perturbations of the active decision boundaries rather than weighting parameter directions according to the
complete-classification-information target \(I_{\mathrm{CC}}\) used by
the Fisher comparator. We therefore interpret the class-specific results as
descriptive evidence that the two design criteria emphasize different
aspects of the multiclass problem, rather than as evidence that
classification-risk acquisition improves every class simultaneously.

Overall, the Landsat application supports two aspects of the proposed
framework. First, the observation-level geometry demonstrates that posterior
uncertainty and classification value can induce materially different
acquisition patterns on the same unlabeled feature pool. Second, the repeated
pilot experiment reveals a systematic distinction between
classification-risk and Fisher-information targeting: the adaptive rule
attained lower mean overall error than Fisher at all three budgets, with the
difference clearly resolved by the paired Monte Carlo comparisons at the
20\% and 30\% budgets. The proposed rule does not uniformly dominate entropy
or margin sampling, particularly at the smallest budget, and the differences
among the targeted procedures become small once labels are less scarce.
These findings support the narrower theoretical claim of the paper: the
value of a label for classification depends not only on the amount of
information it provides, but also on how that information aligns with
parameter directions that perturb the active classification boundaries.

As an additional empirical application,
Supplementary Section~\ref{sec:supp-drybean} reports a
seven-class analysis of the Dry Bean data of \citet{KokluOzkan2020}. The
results provide complementary evidence under a larger feature pool: the
adaptive classification-risk rule attains the smallest mean overall and
balanced classification errors at the most restrictive labeling budget,
while differences among the information-based designs become small as the
labeling budget increases.


\section{Discussion}
\label{sec:discussion}

This paper develops a classification-risk-based principle for allocating a
limited labeling budget in parametric multiclass learning. The central
distinction is between acquiring labels that are globally informative for
parameter estimation and acquiring labels that are informative specifically
for the classification decision. By combining the conditional label
information
\[
J(y)=I_{Z\mid Y}(\theta_0;y)
\]
with the local curvature \(H_R\) of multiclass zero--one excess risk, the
leading classification loss under an acquisition design \(a\) is governed by
\[
\Phi(a)
=
\operatorname{tr}
\{H_RI(a)^{-1}\}.
\]
The corresponding marginal value
\[
\psi_a(y)
=
\operatorname{tr}
\left\{
H_RI(a)^{-1}J(y)I(a)^{-1}
\right\}
\]
therefore measures not simply how much information a label supplies, but how
strongly that information is aligned with parameter directions that perturb
the active Bayes boundary.

This perspective clarifies the relationship between the proposed criterion
and several established approaches to label acquisition. Posterior-uncertainty
rules such as entropy and margin sampling depend only on the ambiguity of the
current class probabilities. The analytic logistic example in
Section~\ref{sec:uncertainty} shows that this information is generally
insufficient: observations with the same posterior uncertainty can have
different classification values, and uncertainty and classification-value
rankings can even be reversed. Fisher-information-based acquisition addresses a different limitation by
accounting for the matrix-valued information contributed by a label. The
particular Fisher comparator used here targets estimation relative to the
complete-classification information matrix \(I_{\mathrm{CC}}\), rather
than directly targeting zero--one classification risk. The distinction
between the two criteria considered here is explicit in
\[
\Phi_{\mathrm F}(a)
=
\operatorname{tr}
\{I_{\mathrm{CC}}I(a)^{-1}\},
\qquad
\Phi(a)
=
\operatorname{tr}
\{H_RI(a)^{-1}\}.
\]
Both are trace-inverse optimal-design criteria; what differs is the target
matrix. The contribution here is therefore not the generic optimization form,
but the derivation of \(H_R\) from multiclass zero--one classification risk
and its use to determine which directions of label information matter for
the classifier.

The active-face decomposition provides a specifically multiclass
interpretation of this principle. Because
\[
H_R=\sum_{k<l}H_{kl},
\]
both the leading excess risk and the marginal acquisition value can be
decomposed over the active pairwise Bayes faces. A candidate observation need
not lie close to a particular face in order to be informative for that face.
Its label may instead improve estimation of parameters that control the
position, orientation, or curvature of the face elsewhere in feature space.
This inferential propagation is one reason why classification-risk-optimal
acquisition need not reduce to conventional boundary uncertainty.

The numerical results are consistent with this interpretation. In the
heterogeneous QDA setting, the entropy and classification-value surfaces have
visibly different geometry, and the adaptive classification-risk rule gives
the smallest mean classification error among the implementable procedures at
each labeling budget considered. When the covariance matrices are made much
more similar, the Bayes boundaries become nearly linear and the Fisher and
classification-risk designs become substantially closer. The differences
among all targeted methods also decrease as the labeling budget increases.
These results should not be interpreted as a universal dominance statement.
Rather, they illustrate that the practical gain from risk-targeted
acquisition depends on the extent to which the available label information
differs in its alignment with classification-relevant parameter directions.

The Landsat application provides a corresponding real-data illustration.
Posterior uncertainty and classification value emphasize different portions
of the observed feature pool and lead to visibly different acquisition
patterns. Across 50 repeated pilot/acquisition experiments, the adaptive
classification-risk design attained lower mean overall error than the
Fisher-information design at all three labeling budgets, with paired
differences clearly favoring the adaptive design at the 20\% and 30\%
budgets. Its performance was broadly comparable with entropy and margin
sampling, and differences among the targeted procedures became small at the
largest budget. The real-data evidence therefore supports a distinction between
classification-risk targeting and the \(I_{\mathrm{CC}}\)-targeted Fisher
comparator rather than a claim of uniform dominance over every
active-learning method. The
supplementary Dry Bean analysis provides complementary evidence in a
different seven-class setting.

The oracle design depends on the unknown feature distribution, conditional
label information, and Bayes-boundary geometry. The two-stage adaptive
construction addresses this dependence by estimating these quantities from a
vanishing pilot fraction. Under the stated regularity and uniform consistency
conditions, the resulting design attains the oracle leading-risk coefficient.
The assumptions are deliberately strong because the optimization ranges over
a large class of acquisition functions. In applications, implementations may
instead restrict the design to a finite candidate pool or to a structured
family of acquisition functions. Such restrictions can substantially simplify
estimation and computation, while retaining the same classification-risk
principle.

Several extensions are natural. First, the present development is local and
parametric. The oracle interpretation also relies on correct specification of
the working classification model. Under model misspecification, the resulting
criterion should be viewed as optimizing the local risk geometry of the
working model rather than the unknown population Bayes rule. Extending the
active-boundary risk geometry to semiparametric or nonparametric classifiers
would require controlling both estimation error and the stability of the
estimated decision boundary. Second, the current theory treats the acquisition
probability as a function of the observed feature vector. More general
sequential designs could also incorporate previously acquired labels, batch
dependence, or other state information. Third, classification costs need not
be symmetric. Replacing zero--one loss by a cost-sensitive loss would change
the relevant decision surfaces and hence the risk-curvature matrix, but the
same information-geometric principle should continue to apply once the
corresponding local risk expansion is available. Finally, richer multiclass
problems may contain nearly coincident or nonregular decision surfaces, for
which the pairwise-face expansion requires additional geometric analysis.

The broader implication is that label acquisition should be matched to the
statistical objective ultimately used to evaluate the classifier. A label is
valuable not merely because its outcome is uncertain and not merely because it
contains Fisher information. Its value depends on whether that information
acts in parameter directions that matter for the downstream decision rule.
For multiclass zero--one classification, those directions are encoded by the
active Bayes-boundary curvature \(H_R\). This yields a direct connection
between optimal label acquisition, Fisher information, and the local geometry
of classification risk.
\section{Conclusion}
\label{sec:conclusion}

We have developed a classification-risk-optimal framework for selective
label acquisition in parametric multiclass learning. The key result is that,
under a controlled acquisition design $a$, the leading expected excess
classification risk is governed by
\[
\operatorname{tr}
\left\{
H_R I(a)^{-1}
\right\},
\]
where $I(a)$ is the Fisher information generated by the observed features
and acquired labels, and $H_R$ is the local curvature of multiclass
zero--one risk over the active Bayes boundary. This leads to the
classification-value function
\[
\psi_a(y)
=
\operatorname{tr}
\left\{
H_R I(a)^{-1} J(y) I(a)^{-1}
\right\},
\]
which quantifies the marginal value of acquiring a label at feature value
$y$. Unlike posterior-uncertainty criteria, this quantity depends on the
matrix-valued information supplied by the label and on its alignment with
parameter directions that affect the decision boundary. Unlike a global
information target such as the complete-classification-information comparator
used here, it weights those directions according to their local contribution
to zero--one classification risk.

The resulting oracle acquisition problem is convex and admits a threshold
characterization, with labels allocated to locations having the largest
marginal classification value. The active-face decomposition further shows
how this value can be attributed to individual pairwise Bayes faces, while
the cost-sensitive extension replaces absolute value by value per unit
labeling cost. A two-stage adaptive procedure was also shown to attain the
oracle leading-risk coefficient under suitable regularity and consistency
conditions.

The Gaussian discriminant specialization provides explicit expressions for
the information and boundary-curvature quantities, making the framework
directly computable for multiclass LDA and QDA. The simulation results
illustrate that classification-risk-based acquisition can differ substantially
from entropy, margin, and the $I_{\mathrm{CC}}$-targeted Fisher comparator,
particularly when labels are scarce and the Bayes geometry is heterogeneous.

The Landsat application provides corresponding real-data evidence:
classification-risk and posterior-uncertainty criteria generate different
acquisition patterns, while the adaptive classification-risk design attains
lower mean error than the $I_{\mathrm{CC}}$-targeted Fisher comparator across
the labeling budgets considered. It does not, however, uniformly outperform
entropy or margin sampling. The additional Dry Bean analysis in the
Supplementary Material provides complementary evidence in a seven-class
setting.

The main message is that the usefulness of a label is determined neither by
posterior uncertainty alone nor by total Fisher information alone. What
matters is whether the information supplied by that label acts in parameter
directions that influence the active multiclass decision boundary. This
provides a direct link between optimal label acquisition, statistical
information, and the local geometry of classification risk.

Code and reproducibility materials for the simulation studies and real-data
applications are available in the
\href{https://github.com/fariborz-setoudeh/classification-risk-optimal-label-acquisition}
{\texttt{GitHub repository}}.


\clearpage

\begin{center}
{\Large\bfseries Supplementary Material for}\\[0.5em]
{\Large\bfseries Classification-Risk-Optimal Label Acquisition}
\end{center}

\vspace{1.5em}


\setcounter{section}{0}
\setcounter{subsection}{0}
\setcounter{equation}{0}
\setcounter{table}{0}
\setcounter{figure}{0}
\setcounter{theorem}{0}

\renewcommand{\thesection}{S\arabic{section}}
\renewcommand{\thesubsection}{\thesection.\arabic{subsection}}
\renewcommand{\theequation}{\thesection.\arabic{equation}}
\renewcommand{\thetable}{S\arabic{table}}
\renewcommand{\thefigure}{S\arabic{figure}}

\renewcommand{\theHsection}{supp.\arabic{section}}
\renewcommand{\theHsubsection}{supp.\arabic{section}.\arabic{subsection}}
\renewcommand{\theHequation}{supp.\arabic{section}.\arabic{equation}}
\renewcommand{\theHtable}{supp.\arabic{table}}
\renewcommand{\theHfigure}{supp.\arabic{figure}}



\section{Information and classification-risk results under controlled acquisition}
\label{sec:supp-acquisition}

This section proves
Propositions~\ref{prop:design-information} and~\ref{prop:design-risk}.

\subsection*{Proof of Proposition~\ref{prop:design-information}}

For one sampling unit, the observed datum is
\[
O=(Y,\Delta,\Delta Z).
\]
Because the acquisition mechanism is controlled and parameter-free,
\[
\Pr(\Delta=1\mid Y=y,Z)=a(y).
\]
The observed-data distribution can therefore be written explicitly as
\[
p_{\theta,a}(y,\Delta=0)
=
p_{\theta}(y)\{1-a(y)\},
\]
and, for \(z\in\{1,\ldots,g\}\),
\[
p_{\theta,a}(y,\Delta=1,Z=z)
=
p_{\theta}(y)\tau_z(y;\theta)a(y).
\]
Since the factors involving \(a\) do not depend on \(\theta\), the score for
\(\theta\) is
\begin{equation}
S_a(O;\theta)
=
S_Y(Y;\theta)
+
\Delta S_{Z\mid Y}(Y,Z;\theta).
\label{eq:supp-design-score}
\end{equation}

For every \(y\),
\begin{align}
E_{\theta}
\left[
S_{Z\mid Y}(Y,Z;\theta)
\mid Y=y
\right]
&=
\sum_{k=1}^{g}
\tau_k(y;\theta)
\nabla_{\theta}
\log\tau_k(y;\theta)
\nonumber\\
&=
\sum_{k=1}^{g}
\nabla_{\theta}\tau_k(y;\theta)
\nonumber\\
&=
0.
\label{eq:supp-conditional-score-zero}
\end{align}
It follows that the two terms in
\eqref{eq:supp-design-score} are orthogonal in expectation. Indeed,
conditional on \(Y\),
\[
E_{\theta}
\left[
\Delta S_{Z\mid Y}\mid Y
\right]
=
a(Y)
E_{\theta}
\left[
S_{Z\mid Y}\mid Y
\right]
=
0,
\]
so that
\[
E_{\theta}
\left[
S_Y
\{\Delta S_{Z\mid Y}\}^{\top}
\right]
=
0.
\]

Consequently,
\begin{align}
I(a;\theta)
&=
E_{\theta}
\left[
S_aS_a^{\top}
\right]
\nonumber\\
&=
I_Y(\theta)
+
E_{\theta}
\left[
\Delta
S_{Z\mid Y}S_{Z\mid Y}^{\top}
\right].
\label{eq:supp-information-expand}
\end{align}
Because \(\Delta^2=\Delta\), and because
\(\Delta\perp Z\mid Y\) under the controlled design,
\begin{align}
E_{\theta}
\left[
\Delta
S_{Z\mid Y}S_{Z\mid Y}^{\top}
\mid Y
\right]
&=
E(\Delta\mid Y)
E_{\theta}
\left[
S_{Z\mid Y}S_{Z\mid Y}^{\top}
\mid Y
\right]
\nonumber\\
&=
a(Y)I_{Z\mid Y}(\theta;Y).
\label{eq:supp-label-info}
\end{align}
Substitution into \eqref{eq:supp-information-expand} gives
\[
I(a;\theta)
=
I_Y(\theta)
+
E_{\theta}
\left[
a(Y)I_{Z\mid Y}(\theta;Y)
\right],
\]
which proves \eqref{eq:design-information}.

Under complete classification, \(\Delta=1\) almost surely, and therefore
\[
I_{\mathrm{CC}}(\theta)
=
I_Y(\theta)
+
E_{\theta}
\left[
I_{Z\mid Y}(\theta;Y)
\right].
\]
Subtracting
\[
E_{\theta}
\left[
\{1-a(Y)\}I_{Z\mid Y}(\theta;Y)
\right]
\]
from this expression gives
\[
I(a;\theta)
=
I_{\mathrm{CC}}(\theta)
-
E_{\theta}
\left[
\{1-a(Y)\}I_{Z\mid Y}(\theta;Y)
\right].
\]
This proves Proposition~\ref{prop:design-information}.
\(\square\)

\subsection*{Proof of Proposition~\ref{prop:design-risk}}

Write
\[
h_n=\widehat{\theta}_a-\theta_0.
\]
By the assumed regular asymptotic normality,
\[
\sqrt{n}h_n
\overset{d}{\longrightarrow}
Z_a,
\qquad
Z_a\sim
N\!\left(
0,
I(a;\theta_0)^{-1}
\right).
\]

The local multiclass excess-risk expansion gives
\[
R(\theta_0+h_n)-R^*
=
\frac{1}{2}
h_n^{\top}H_Rh_n
+
o_p(n^{-1}),
\]
because \(h_n=O_p(n^{-1/2})\). Multiplying by \(n\),
\begin{equation}
n\{R(\widehat{\theta}_a)-R^*\}
=
\frac{1}{2}
(\sqrt{n}h_n)^{\top}
H_R
(\sqrt{n}h_n)
+
o_p(1).
\label{eq:supp-risk-limit}
\end{equation}

Under the stated moment condition, the sequence on the left-hand side of
\eqref{eq:supp-risk-limit} is uniformly integrable. Hence convergence in
distribution together with uniform integrability yields
\begin{align}
n
\left[
E_{\theta_0}
\{R(\widehat{\theta}_a)\}
-
R^*
\right]
&\longrightarrow
\frac{1}{2}
E
\left[
Z_a^{\top}H_RZ_a
\right]
\nonumber\\
&=
\frac{1}{2}
\operatorname{tr}
\left[
H_R
E(Z_aZ_a^{\top})
\right]
\nonumber\\
&=
\frac{1}{2}
\operatorname{tr}
\left\{
H_RI(a;\theta_0)^{-1}
\right\}.
\end{align}
Therefore,
\[
E_{\theta_0}
\{R(\widehat{\theta}_a)\}
-
R^*
=
\frac{1}{2n}
\operatorname{tr}
\left\{
H_RI(a;\theta_0)^{-1}
\right\}
+
o(n^{-1}),
\]
which proves Proposition~\ref{prop:design-risk}.
\(\square\)

\section{Proof of the oracle acquisition result}
\label{sec:supp-oracle-design}

This section proves Theorem~\ref{thm:oracle-design}
and Corollary~\ref{cor:bang-bang-design}. Expectations are taken under
\(\theta_0\), and we write
\[
J(Y)=I_{Z\mid Y}(\theta_0;Y).
\]

We first record two elementary properties of the design criterion.

\subsection*{Budget saturation}

Let \(a\) and \(\widetilde a\) be two acquisition designs satisfying
\[
0\leq a(y)\leq\widetilde a(y)\leq1
\]
almost everywhere. Then
\[
I(\widetilde a)-I(a)
=
E\!\left[
\{\widetilde a(Y)-a(Y)\}J(Y)
\right]
\succeq0,
\]
because \(J(Y)\succeq0\). Whenever both information matrices are positive
definite,
\[
I(\widetilde a)^{-1}
\preceq
I(a)^{-1}.
\]
Since \(H_R\succeq0\),
\[
\operatorname{tr}
\{H_RI(\widetilde a)^{-1}\}
\leq
\operatorname{tr}
\{H_RI(a)^{-1}\}.
\]
Thus allocating additional labeling probability cannot increase the
criterion.

Now suppose that \(E\{a(Y)\}<\rho\). Since \(\rho<1\),
\[
E\{1-a(Y)\}>0.
\]
Define
\[
\widetilde a(y)
=
a(y)+t\{1-a(y)\},
\qquad
t
=
\frac{\rho-E\{a(Y)\}}
     {E\{1-a(Y)\}}.
\]
Then \(0<t\leq1\), \(a\leq\widetilde a\leq1\), and
\[
E\{\widetilde a(Y)\}=\rho.
\]
The preceding monotonicity argument therefore gives
\[
\Phi(\widetilde a)\leq\Phi(a).
\]
Hence the infimum under the constraint
\(E\{a(Y)\}\leq\rho\) equals the infimum over
\(\mathcal A_\rho\).

\subsection*{Convexity and existence}

For \(a_0,a_1\in\mathcal A_\rho\) and \(t\in[0,1]\), define
\[
a_t=(1-t)a_0+ta_1.
\]
Then \(a_t\in\mathcal A_\rho\) and
\[
I(a_t)
=
(1-t)I(a_0)+tI(a_1).
\]
The map
\[
M\longmapsto \operatorname{tr}(H_RM^{-1})
\]
is convex on the cone of positive-definite matrices whenever
\(H_R\succeq0\). Therefore
\[
\Phi(a_t)
\leq
(1-t)\Phi(a_0)+t\Phi(a_1),
\]
which proves convexity of \(\Phi\).

For completeness, the same conclusion follows directly by differentiating
along the segment. Let
\[
B=I(a_1)-I(a_0)
\]
and \(M_t=I(a_t)\). Then
\[
\frac{d}{dt}M_t^{-1}
=
-M_t^{-1}BM_t^{-1},
\]
and hence
\[
\frac{d^2}{dt^2}
\operatorname{tr}(H_RM_t^{-1})
=
2\operatorname{tr}
\left(
H_RM_t^{-1}BM_t^{-1}BM_t^{-1}
\right).
\]
Writing
\[
C_t=M_t^{-1/2}BM_t^{-1/2},
\qquad
K_t=M_t^{-1/2}H_RM_t^{-1/2},
\]
gives
\[
\operatorname{tr}
\left(
H_RM_t^{-1}BM_t^{-1}BM_t^{-1}
\right)
=
\operatorname{tr}(K_tC_t^2)
\geq0,
\]
because \(K_t\succeq0\) and \(C_t^2\succeq0\).

To establish existence, regard \(\mathcal A_\rho\) as a subset of
\(L^\infty(P_{\theta_0})\). The constraints
\[
0\leq a\leq1,
\qquad
E(a)=\rho,
\]
define a weak-\(*\) compact set. Since every entry of \(J(Y)\) is integrable,
the map
\[
a\longmapsto I(a)
=
I_Y+E\{a(Y)J(Y)\}
\]
is weak-\(*\) continuous. By the assumed uniform nonsingularity of
\(I(a)\) over \(\mathcal A_\rho\),
\[
a\longmapsto
\operatorname{tr}\{H_RI(a)^{-1}\}
\]
is continuous on this set. It therefore attains its minimum.

This proves part~(i) of Theorem~\ref{thm:oracle-design}.

\subsection*{Directional derivative}

Fix \(a\in\mathcal A_\rho\), and let \(h\) be a bounded measurable
perturbation such that \(a+th\) remains admissible for all sufficiently small
\(t\geq0\). Put
\[
B_h=E\{h(Y)J(Y)\}.
\]
Then
\[
I(a+th)=I(a)+tB_h.
\]
The right derivative along the feasible direction gives
\[
\left.
\frac{d}{dt}
I(a+th)^{-1}
\right|_{t=0}
=
-I(a)^{-1}B_hI(a)^{-1}.
\]
Consequently,
\begin{align}
D\Phi(a)[h]
&=
-\operatorname{tr}
\left\{
H_RI(a)^{-1}
B_h
I(a)^{-1}
\right\}
\nonumber\\
&=
-
E
\left[
h(Y)
\operatorname{tr}
\left\{
H_RI(a)^{-1}
J(Y)
I(a)^{-1}
\right\}
\right]
\nonumber\\
&=
-E\{h(Y)\psi_a(Y)\}.
\label{eq:supp-directional-derivative}
\end{align}
This proves part~(ii).

\subsection*{Equivalence condition}

Because \(\Phi\) is convex, a feasible design \(a^\star\) minimizes
\(\Phi\) over the convex set \(\mathcal A_\rho\) if and only if
\begin{equation}
D\Phi(a^\star)[a-a^\star]\geq0
\qquad
\text{for every }a\in\mathcal A_\rho.
\label{eq:supp-first-order-condition}
\end{equation}
Using \eqref{eq:supp-directional-derivative},
\[
D\Phi(a^\star)[a-a^\star]
=
-
E
\left[
\{a(Y)-a^\star(Y)\}
\psi_{a^\star}(Y)
\right].
\]
Thus \eqref{eq:supp-first-order-condition} is equivalent to
\[
E
\left[
a(Y)\psi_{a^\star}(Y)
\right]
\leq
E
\left[
a^\star(Y)\psi_{a^\star}(Y)
\right]
\]
for every \(a\in\mathcal A_\rho\). This proves part~(iii).

The result has a useful interpretation. Conditional on
\(\psi_{a^\star}\), the optimal design \(a^\star\) solves the linear problem
\begin{equation}
\max_{a\in\mathcal A_\rho}
E\{a(Y)\psi_{a^\star}(Y)\}.
\label{eq:supp-linear-problem}
\end{equation}

\subsection*{Threshold characterization}

Let
\[
\psi^\star(Y)=\psi_{a^\star}(Y).
\]
Since \(J(Y)\) is integrable and \(I(a^\star)^{-1}\) is bounded,
\(\psi^\star(Y)\) is integrable.

Choose \(\lambda^\star\) satisfying
\begin{equation}
\Pr\{\psi^\star(Y)>\lambda^\star\}
\leq
\rho
\leq
\Pr\{\psi^\star(Y)\geq\lambda^\star\}.
\label{eq:supp-threshold-quantile}
\end{equation}
Such a value exists by the usual quantile construction.

The maximizers of \eqref{eq:supp-linear-problem} allocate all available
weight first to locations having the largest values of \(\psi^\star\).
More precisely, any maximizer must satisfy, almost surely,
\[
a^\star(Y)=1
\quad\text{on}\quad
\{\psi^\star(Y)>\lambda^\star\},
\]
and
\[
a^\star(Y)=0
\quad\text{on}\quad
\{\psi^\star(Y)<\lambda^\star\}.
\]
On the threshold set
\[
\{\psi^\star(Y)=\lambda^\star\},
\]
any allocation in \([0,1]\) that makes
\(E\{a^\star(Y)\}=\rho\) has the same value in the linear problem.

For completeness, this follows directly from an exchange argument. Suppose
there exist measurable sets \(A\) and \(B\), each of positive probability,
such that
\[
\psi^\star(y_1)<\psi^\star(y_2)
\qquad
(y_1\in A,\ y_2\in B),
\]
while \(a^\star\) assigns positive acquisition probability on \(A\) and
leaves positive unused acquisition capacity on \(B\). Transferring a
sufficiently small amount of acquisition probability from \(A\) to \(B\)
preserves the budget but strictly increases
\[
E\{a(Y)\psi^\star(Y)\},
\]
contradicting optimality in \eqref{eq:supp-linear-problem}. Hence lower-value
locations cannot receive acquisition probability while higher-value
locations remain unsaturated.

It follows that there is a threshold \(\lambda^\star\) such that
\[
\begin{cases}
\psi_{a^\star}(Y)\leq\lambda^\star,
&
a^\star(Y)=0,
\\[3pt]
\psi_{a^\star}(Y)=\lambda^\star,
&
0<a^\star(Y)<1,
\\[3pt]
\psi_{a^\star}(Y)\geq\lambda^\star,
&
a^\star(Y)=1,
\end{cases}
\]
almost surely. Conversely, any feasible design satisfying this threshold
condition maximizes \eqref{eq:supp-linear-problem}; by part~(iii), it is
therefore a global minimizer of \(\Phi\).

This proves part~(iv) and completes the proof of
Theorem~\ref{thm:oracle-design}.
\(\square\)

\subsection*{Proof of Corollary~\ref{cor:bang-bang-design}}

If
\[
\Pr\{\psi_{a^\star}(Y)=\lambda^\star\}=0,
\]
the threshold set has probability zero. Part~(iv) of Theorem~\ref{thm:oracle-design} therefore implies
\[
a^\star(Y)
=
\mathbf 1
\{\psi_{a^\star}(Y)>\lambda^\star\}
\]
almost surely. Since \(a^\star\in\mathcal A_\rho\),
\[
\rho
=
E\{a^\star(Y)\}
=
\Pr
\{\psi_{a^\star}(Y)>\lambda^\star\}.
\]
This proves the result.
\(\square\)

\section{Posterior uncertainty and classification value}
\label{sec:supp-uncertainty}

This section proves Proposition~\ref{prop:uncertainty-counterexample}.

Let
\[
Y=(Y_1,Y_2)^{\top}\sim N_2(0,I_2),
\qquad
\boldsymbol x=(1,Y_1,Y_2)^{\top},
\]
and consider the conditional model
\[
\Pr_{\boldsymbol\beta}(Z=1\mid Y)
=
\sigma(\boldsymbol x^{\top}\boldsymbol\beta)
\]
at
\[
\boldsymbol\beta_0=(0,b,0)^{\top},
\qquad b>0.
\]
Throughout the proof, write
\[
\eta(y)=\sigma(by_1),
\qquad
v(by_1)=\eta(y)\{1-\eta(y)\}.
\]

The marginal distribution of \(Y\) is parameter-free, so the feature
information \(I_Y\) is zero. The conditional Bernoulli score is
\[
S_{Z\mid Y}
=
\{Z-\eta(Y)\}\boldsymbol x,
\]
and therefore
\begin{equation}
J(y)
=
I_{Z\mid Y}(\boldsymbol\beta_0;y)
=
v(by_1)\boldsymbol x\boldsymbol x^{\top}.
\label{eq:supp-logistic-J}
\end{equation}

We first derive the classification-risk curvature matrix \(H_R\).
Let \(p_Y(y)\) denote the \(N_2(0,I_2)\) density. The difference between the
two prior-weighted class densities is
\[
g(y;\boldsymbol\beta)
=
p_Y(y)
\left[
2\sigma(\boldsymbol x^{\top}\boldsymbol\beta)-1
\right].
\]
At \(\boldsymbol\beta_0\), the Bayes boundary is
\[
\mathcal F=\{y:y_1=0\}.
\]

For \(s=(0,s_2)^{\top}\in\mathcal F\),
\[
\sigma(\boldsymbol x^{\top}\boldsymbol\beta_0)
=
\frac12.
\]
Differentiating \(g\) with respect to \(\boldsymbol\beta\) gives
\[
\nabla_{\boldsymbol\beta}
g(s;\boldsymbol\beta_0)
=
2p_Y(s)
\sigma(0)\{1-\sigma(0)\}
\begin{pmatrix}
1\\
0\\
s_2
\end{pmatrix}
=
\frac{p_Y(s)}{2}
\begin{pmatrix}
1\\
0\\
s_2
\end{pmatrix}.
\]
Similarly,
\[
\nabla_y g(s;\boldsymbol\beta_0)
=
\frac{p_Y(s)}{2}
\begin{pmatrix}
b\\
0
\end{pmatrix},
\]
because the term involving \(\nabla_y p_Y(s)\) is multiplied by
\(2\sigma(0)-1=0\). Hence
\[
\left\|
\nabla_y g(s;\boldsymbol\beta_0)
\right\|
=
\frac{bp_Y(s)}{2}.
\]

Using the binary specialization of the Bayes-boundary curvature formula,
\begin{align}
H_R
&=
\int_{\mathcal F}
\frac{
\nabla_{\boldsymbol\beta}g(s;\boldsymbol\beta_0)
\nabla_{\boldsymbol\beta}g(s;\boldsymbol\beta_0)^{\top}
}{
\|\nabla_y g(s;\boldsymbol\beta_0)\|
}
\,dS(s)
\nonumber\\
&=
\frac{1}{2b}
\int_{-\infty}^{\infty}
p_Y(0,s_2)
\begin{pmatrix}
1\\
0\\
s_2
\end{pmatrix}
\begin{pmatrix}
1&0&s_2
\end{pmatrix}
\,ds_2.
\label{eq:supp-H-integral}
\end{align}
Since
\[
p_Y(0,s_2)=\phi(0)\phi(s_2),
\]
symmetry and the first two moments of a standard normal random variable give
\begin{equation}
H_R
=
\frac{\phi(0)}{2b}
\begin{pmatrix}
1&0&0\\
0&0&0\\
0&0&1
\end{pmatrix}.
\label{eq:supp-logistic-H}
\end{equation}

Now consider the uniform acquisition design
\[
a_\rho(y)\equiv\rho.
\]
By Proposition~\ref{prop:design-information} and
\eqref{eq:supp-logistic-J},
\[
I(a_\rho)
=
\rho E
\left[
v(bY_1)\boldsymbol x\boldsymbol x^{\top}
\right].
\]
Define
\[
m_0
=
E\{v(bY_1)\},
\qquad
m_2
=
E\{Y_1^2v(bY_1)\}.
\]
The function \(v(bt)\) is even in \(t\), and \(Y_1\) and \(Y_2\) are
independent centered standard normal variables. All off-diagonal entries
therefore vanish, and
\begin{equation}
I(a_\rho)
=
\rho
\begin{pmatrix}
m_0&0&0\\
0&m_2&0\\
0&0&m_0
\end{pmatrix}.
\label{eq:supp-uniform-information}
\end{equation}
Both \(m_0\) and \(m_2\) are strictly positive, so this matrix is positive
definite.

Its inverse is
\[
I(a_\rho)^{-1}
=
\frac{1}{\rho}
\begin{pmatrix}
m_0^{-1}&0&0\\
0&m_2^{-1}&0\\
0&0&m_0^{-1}
\end{pmatrix}.
\]
Combining this expression with
\eqref{eq:supp-logistic-H} gives
\begin{equation}
I(a_\rho)^{-1}
H_R
I(a_\rho)^{-1}
=
\frac{\phi(0)}
{2b\rho^2m_0^2}
\begin{pmatrix}
1&0&0\\
0&0&0\\
0&0&1
\end{pmatrix}.
\label{eq:supp-G-logistic}
\end{equation}

Using cyclic invariance of the trace together with
\eqref{eq:supp-logistic-J},
\begin{align}
\psi_{a_\rho}(y)
&=
\operatorname{tr}
\left[
I(a_\rho)^{-1}
H_R
I(a_\rho)^{-1}
J(y)
\right]
\nonumber\\
&=
\frac{\phi(0)}
{2b\rho^2m_0^2}
v(by_1)
\operatorname{tr}
\left[
\begin{pmatrix}
1&0&0\\
0&0&0\\
0&0&1
\end{pmatrix}
\boldsymbol x\boldsymbol x^{\top}
\right]
\nonumber\\
&=
\frac{\phi(0)}
{2b\rho^2m_0^2}
v(by_1)(1+y_2^2).
\label{eq:supp-classification-value}
\end{align}
This proves \eqref{eq:logistic-classification-value}.

Posterior entropy is
\[
U_{\mathrm{ent}}(y)
=
h\{\eta(y)\}
=
h\{\sigma(by_1)\},
\]
and hence is independent of \(y_2\). For fixed \(y_1\),
\[
\psi_{a_\rho}(y_1,y_2)
=
C_{\rho,b}v(by_1)(1+y_2^2),
\]
so two points with the same \(y_1\) but different values of \(y_2^2\) have
identical posterior entropy and different classification values. This proves
part~(i).

To prove part~(ii), take
\[
y^{(1)}=(0,0)^{\top}.
\]
Since \(\sigma(0)=1/2\), this point has maximum binary posterior entropy:
\[
U_{\mathrm{ent}}\{y^{(1)}\}=\log 2.
\]
Moreover,
\[
\psi_{a_\rho}\{y^{(1)}\}
=
\frac{C_{\rho,b}}{4}.
\]

Now fix any \(\delta\neq0\) and let
\[
y^{(2)}=(\delta,M)^{\top}.
\]
Because
\[
\sigma(b\delta)\neq\frac12,
\]
strict concavity of binary entropy implies
\[
U_{\mathrm{ent}}\{y^{(2)}\}<\log2
=
U_{\mathrm{ent}}\{y^{(1)}\}.
\]
On the other hand,
\[
\psi_{a_\rho}\{y^{(2)}\}
=
C_{\rho,b}
v(b\delta)(1+M^2).
\]
Since \(v(b\delta)>0\), choose \(M\) sufficiently large that
\[
v(b\delta)(1+M^2)>\frac14.
\]
Equivalently, it is enough to take
\[
M^2>
\frac{1}{4v(b\delta)}-1.
\]
Then
\[
\psi_{a_\rho}\{y^{(2)}\}
>
\frac{C_{\rho,b}}{4}
=
\psi_{a_\rho}\{y^{(1)}\}.
\]
Thus
\[
U_{\mathrm{ent}}\{y^{(1)}\}
>
U_{\mathrm{ent}}\{y^{(2)}\},
\qquad
\psi_{a_\rho}\{y^{(1)}\}
<
\psi_{a_\rho}\{y^{(2)}\},
\]
which proves the strict ranking reversal in
\eqref{eq:ranking-reversal}.

This completes the proof of
Proposition~\ref{prop:uncertainty-counterexample}.
\(\square\)
\section{Face-specific and cost-sensitive acquisition results}
\label{sec:supp-face-specific}

This section proves Proposition~\ref{prop:face-decomposition}
and Corollary~\ref{cor:cost-sensitive-design}. All expectations are taken under
\(\theta_0\).

\subsection*{Proof of Proposition~\ref{prop:face-decomposition}}

Recall that
\[
H_R
=
\sum_{1\leq k<l\leq g}H_{kl}.
\]
By linearity of the trace,
\begin{align}
\Phi(a)
&=
\operatorname{tr}
\left\{
H_RI(a)^{-1}
\right\}
\nonumber\\
&=
\operatorname{tr}
\left\{
\left(
\sum_{k<l}H_{kl}
\right)
I(a)^{-1}
\right\}
\nonumber\\
&=
\sum_{k<l}
\operatorname{tr}
\left\{
H_{kl}I(a)^{-1}
\right\}
\nonumber\\
&=
\sum_{k<l}\Phi_{kl}(a),
\end{align}
which proves \eqref{eq:criterion-face-decomposition}.

Similarly,
\begin{align}
\psi_a(y)
&=
\operatorname{tr}
\left\{
H_RI(a)^{-1}
J(y)
I(a)^{-1}
\right\}
\nonumber\\
&=
\sum_{k<l}
\operatorname{tr}
\left\{
H_{kl}I(a)^{-1}
J(y)
I(a)^{-1}
\right\}
\nonumber\\
&=
\sum_{k<l}\psi_{kl,a}(y),
\end{align}
which proves \eqref{eq:value-face-decomposition}.

Since
\[
H_{kl}\succeq0,
\qquad
I(a)^{-1}\succ0,
\]
we have
\[
I(a)^{-1/2}
H_{kl}
I(a)^{-1/2}
\succeq0.
\]
Therefore
\[
\Phi_{kl}(a)
=
\operatorname{tr}
\left[
I(a)^{-1/2}
H_{kl}
I(a)^{-1/2}
\right]
\geq0.
\]

Likewise,
\[
G_{kl,a}
=
I(a)^{-1}
H_{kl}
I(a)^{-1}
\succeq0,
\]
and \(J(y)\succeq0\). Hence
\[
\psi_{kl,a}(y)
=
\operatorname{tr}
\{G_{kl,a}J(y)\}
\geq0,
\]
because the trace of the product of two positive-semidefinite matrices is
nonnegative.

It remains to establish the directional derivative. Let \(h\) be an
admissible perturbation and write
\[
B_h=E\{h(Y)J(Y)\}.
\]
As in Supplementary Section~\ref{sec:supp-oracle-design},
\[
\left.
\frac{d}{dt}
I(a+th)^{-1}
\right|_{t=0}
=
-
I(a)^{-1}B_hI(a)^{-1}.
\]
Consequently,
\begin{align}
D\Phi_{kl}(a)[h]
&=
-
\operatorname{tr}
\left\{
H_{kl}
I(a)^{-1}
B_h
I(a)^{-1}
\right\}
\nonumber\\
&=
-
E
\left[
h(Y)
\operatorname{tr}
\left\{
H_{kl}
I(a)^{-1}
J(Y)
I(a)^{-1}
\right\}
\right]
\nonumber\\
&=
-
E
\left[
h(Y)\psi_{kl,a}(Y)
\right].
\end{align}
This proves \eqref{eq:face-directional-derivative} and completes the proof of
Proposition~\ref{prop:face-decomposition}.
\(\square\)

\subsection*{Proof of Corollary~\ref{cor:cost-sensitive-design}}

Define
\[
\mathcal A_B^c
=
\left\{
a:\mathcal Y\to[0,1]:
a\ \text{measurable},\
E\{c(Y)a(Y)\}\leq B
\right\}.
\]

Suppose that
\[
E\{c(Y)a(Y)\}<B.
\]
Because \(B<E\{c(Y)\}\),
\[
E[c(Y)\{1-a(Y)\}]>0.
\]
Define
\[
\widetilde a(y)
=
a(y)+t\{1-a(y)\},
\qquad
t
=
\frac{B-E\{c(Y)a(Y)\}}
     {E[c(Y)\{1-a(Y)\}]}.
\]
Then \(0<t\leq1\), \(a\leq\widetilde a\leq1\), and
\[
E\{c(Y)\widetilde a(Y)\}=B.
\]
Since \(J(Y)\succeq0\),
\[
I(\widetilde a)\succeq I(a),
\]
and therefore
\[
\Phi(\widetilde a)\leq\Phi(a).
\]
Hence an oracle design can always be chosen to satisfy the budget with
equality.

Existence follows by the same weak-\(*\) compactness argument used in
Supplementary Section~\ref{sec:supp-oracle-design}. We may therefore restrict attention to
\[
\mathcal A_{B,=}^{c}
=
\left\{
a:\mathcal Y\to[0,1]:
E\{c(Y)a(Y)\}=B
\right\}.
\]

Let \(a_B^\star\) be an oracle design. Convexity of \(\Phi\) gives the
first-order condition
\[
D\Phi(a_B^\star)[a-a_B^\star]\geq0
\qquad
\text{for every }
a\in\mathcal A_{B,=}^{c}.
\]
Using the directional-derivative calculation from
Theorem~\ref{thm:oracle-design},
\[
D\Phi(a_B^\star)[a-a_B^\star]
=
-
E
\left[
\{a(Y)-a_B^\star(Y)\}
\psi_{a_B^\star}(Y)
\right].
\]
Thus \(a_B^\star\) maximizes
\begin{equation}
E
\left[
a(Y)\psi_{a_B^\star}(Y)
\right]
\label{eq:supp-cost-linear}
\end{equation}
subject to
\[
0\leq a(Y)\leq1,
\qquad
E\{c(Y)a(Y)\}=B.
\]

Since \(c(Y)>0\), define
\[
\chi^\star(Y)
=
\frac{\psi_{a_B^\star}(Y)}{c(Y)}.
\]
Then
\[
E
\left[
a(Y)\psi_{a_B^\star}(Y)
\right]
=
E
\left[
c(Y)a(Y)\chi^\star(Y)
\right].
\]
The linear problem therefore allocates the available cost first to locations
having the largest values of \(\chi^\star\).

More formally, suppose that there exist measurable sets \(A\) and \(D\), of
positive probability, on which
\[
\chi^\star(y_A)<\chi^\star(y_D),
\]
while \(a_B^\star\) uses positive acquisition capacity on \(A\) and leaves
unused capacity on \(D\). A sufficiently small cost-preserving transfer of
acquisition probability from \(A\) to \(D\) strictly increases
\eqref{eq:supp-cost-linear}, contradicting optimality.

Consequently, there exists
\(\lambda_B^\star\) such that, almost surely,
\[
\begin{cases}
\chi^\star(Y)\leq\lambda_B^\star,
&
a_B^\star(Y)=0,
\\[3pt]
\chi^\star(Y)=\lambda_B^\star,
&
0<a_B^\star(Y)<1,
\\[3pt]
\chi^\star(Y)\geq\lambda_B^\star,
&
a_B^\star(Y)=1.
\end{cases}
\]

If
\[
\Pr
\{\chi^\star(Y)=\lambda_B^\star\}=0,
\]
the intermediate case occurs only on a null set. Hence
\[
a_B^\star(Y)
=
\mathbf 1
\{\chi^\star(Y)>\lambda_B^\star\}
\]
almost surely. The budget constraint then implies
\[
E
\left[
c(Y)
\mathbf 1
\{\chi^\star(Y)>\lambda_B^\star\}
\right]
=
B.
\]
This proves Corollary~\ref{cor:cost-sensitive-design}.
\(\square\)
\section{Adaptive acquisition and oracle classification risk}
\label{sec:supp-adaptive}

This section proves Theorems~\ref{thm:adaptive-oracle}
and~\ref{thm:adaptive-risk}.

Throughout, write
\[
b(a)=E_{\theta_0}\{a(Y)\}
\]
and
\[
\Phi_\rho^\star
=
\inf_{0\leq a\leq1,\;b(a)\leq\rho}
\Phi(a).
\]

We first record a uniform continuity fact used in the plug-in argument.

\subsection*{Uniform convergence of the plug-in criterion}

Suppose
\[
\sup_{0\leq a\leq1}
\|
\widehat I_m(a)-I(a)
\|
\overset{p}{\longrightarrow}0,
\qquad
\|
\widehat H_m-H_R
\|
\overset{p}{\longrightarrow}0,
\]
and suppose there exists \(\kappa>0\) such that, with probability tending to
one,
\[
\lambda_{\min}\{I(a)\}\geq\kappa,
\qquad
\lambda_{\min}\{\widehat I_m(a)\}\geq\kappa
\]
uniformly over \(0\leq a\leq1\).

Then
\[
\sup_{0\leq a\leq1}
\left|
\widehat\Phi_m(a)-\Phi(a)
\right|
\overset{p}{\longrightarrow}0.
\]

Indeed,
\begin{align}
&
\left|
\operatorname{tr}
\{\widehat H_m\widehat I_m(a)^{-1}\}
-
\operatorname{tr}
\{H_RI(a)^{-1}\}
\right|
\nonumber\\
&\quad\leq
\left|
\operatorname{tr}
\left[
(\widehat H_m-H_R)
\widehat I_m(a)^{-1}
\right]
\right|
+
\left|
\operatorname{tr}
\left[
H_R
\{\widehat I_m(a)^{-1}-I(a)^{-1}\}
\right]
\right|.
\label{eq:supp-plugin-bound}
\end{align}
The first term converges uniformly to zero by boundedness of the inverse
information matrices. For the second, the resolvent identity gives
\[
\widehat I_m(a)^{-1}-I(a)^{-1}
=
\widehat I_m(a)^{-1}
\{I(a)-\widehat I_m(a)\}
I(a)^{-1},
\]
and therefore
\[
\sup_a
\|
\widehat I_m(a)^{-1}-I(a)^{-1}
\|
\overset{p}{\longrightarrow}0.
\]
Substitution into \eqref{eq:supp-plugin-bound} establishes the claim.

The rate condition in
\eqref{eq:uniform-criterion-consistency} follows whenever
the preceding convergences hold at a rate \(o_p(\varepsilon_m)\).

\subsection*{Proof of Theorem~\ref{thm:adaptive-oracle}}

Let
\[
r_m
=
\sup_{0\leq a\leq1}
\left|
\widehat b_m(a)-b(a)
\right|.
\]
By assumption,
\[
r_m=o_p(\varepsilon_m).
\]
Hence
\[
\Pr(r_m<\varepsilon_m)\longrightarrow1.
\]

On the event \(r_m<\varepsilon_m\), feasibility of
\(\widehat a_m\) for the fitted problem implies
\[
b(\widehat a_m)
\leq
\widehat b_m(\widehat a_m)+r_m
\leq
\rho-\varepsilon_m+r_m
\leq\rho.
\]
This proves \eqref{eq:adaptive-feasibility}.

We next establish optimality of the achieved population criterion. By the budget-saturation result in Supplementary Section~\ref{sec:supp-oracle-design}, choose an oracle design
\(a^\star\) satisfying
\[
b(a^\star)=\rho,
\qquad
\Phi(a^\star)=\Phi_\rho^\star.
\]

For sufficiently large \(m\), define
\begin{equation}
a_m^\circ
=
\left(
1-\frac{3\varepsilon_m}{\rho}
\right)a^\star.
\label{eq:supp-interior-comparator}
\end{equation}
Since \(\varepsilon_m\downarrow0\),
\[
0\leq a_m^\circ\leq1
\]
for all sufficiently large \(m\), and
\[
b(a_m^\circ)
=
\rho-3\varepsilon_m.
\]

On the event \(r_m<\varepsilon_m\),
\[
\widehat b_m(a_m^\circ)
\leq
b(a_m^\circ)+r_m
<
\rho-2\varepsilon_m
<
\rho-\varepsilon_m.
\]
Thus \(a_m^\circ\) is feasible for the fitted optimization problem with
probability tending to one.

Let
\[
q_m
=
\sup_{0\leq a\leq1}
\left|
\widehat\Phi_m(a)-\Phi(a)
\right|.
\]
By assumption,
\[
q_m=o_p(\varepsilon_m),
\]
and in particular \(q_m=o_p(1)\).

If \(\widehat a_m\) is an exact minimizer, then on the event that
\(a_m^\circ\) is fitted-feasible,
\[
\widehat\Phi_m(\widehat a_m)
\leq
\widehat\Phi_m(a_m^\circ).
\]
For an \(o_p(1)\)-approximate minimizer, the right-hand side acquires only an
additional \(o_p(1)\) term. Hence
\begin{align}
\Phi(\widehat a_m)
&\leq
\widehat\Phi_m(\widehat a_m)+q_m
\nonumber\\
&\leq
\widehat\Phi_m(a_m^\circ)+q_m+o_p(1)
\nonumber\\
&\leq
\Phi(a_m^\circ)+2q_m+o_p(1).
\label{eq:supp-upper-oracle}
\end{align}

Now
\[
I(a_m^\circ)
=
I_Y+
\left(
1-\frac{3\varepsilon_m}{\rho}
\right)
E\{a^\star(Y)J(Y)\}.
\]
Therefore
\[
I(a_m^\circ)\longrightarrow I(a^\star),
\]
and continuity of matrix inversion on the positive-definite cone gives
\[
\Phi(a_m^\circ)\longrightarrow\Phi(a^\star)
=
\Phi_\rho^\star.
\]
Equation~\eqref{eq:supp-upper-oracle} consequently yields
\[
\limsup_{m\to\infty}
\Phi(\widehat a_m)
\leq
\Phi_\rho^\star
\]
in probability.

On the other hand, by \eqref{eq:adaptive-feasibility},
\(\widehat a_m\) is population-feasible with probability tending to one.
Therefore
\[
\Phi(\widehat a_m)\geq\Phi_\rho^\star
\]
with probability tending to one. Combining the upper and lower bounds gives
\[
\Phi(\widehat a_m)
\overset{p}{\longrightarrow}
\Phi_\rho^\star.
\]
This proves Theorem~\ref{thm:adaptive-oracle}.
\(\square\)

\subsection*{Proof of Theorem~\ref{thm:adaptive-risk}}

Let \(\mathcal F_{m_n}\) denote the sigma-field generated by the pilot
experiment and let
\[
N_n=n-m_n.
\]
Conditional on \(\mathcal F_{m_n}\), the fitted acquisition function
\(\widehat a_{m_n}\) is fixed. The \(N_n\) second-stage observations are
independent of the pilot sample and satisfy
\[
\Delta_j\mid Y_j,\mathcal F_{m_n}
\sim
\operatorname{Bernoulli}
\{\widehat a_{m_n}(Y_j)\}.
\]
Therefore, by Proposition~\ref{prop:design-information}, the Fisher
information contributed by one second-stage observation, conditionally on
the pilot stage, is
\[
I(\widehat a_{m_n}).
\]

By the assumed conditional regularity and efficiency,
\[
\sqrt{N_n}
(\widehat\theta_n^{(2)}-\theta_0)
\mid\mathcal F_{m_n}
\]
is asymptotically centered normal with covariance
\[
I(\widehat a_{m_n})^{-1},
\]
uniformly over the fitted sequence of designs.

Applying the quadratic excess-risk expansion conditionally on
\(\mathcal F_{m_n}\), together with the assumed uniform moment condition,
gives
\begin{equation}
2N_n
E_{\theta_0}
\left[
R(\widehat\theta_n^{(2)})-R^*
\mid
\mathcal F_{m_n}
\right]
=
\operatorname{tr}
\left\{
H_RI(\widehat a_{m_n})^{-1}
\right\}
+
o_p(1).
\label{eq:supp-conditional-risk}
\end{equation}
By definition,
\[
\operatorname{tr}
\left\{
H_RI(\widehat a_{m_n})^{-1}
\right\}
=
\Phi(\widehat a_{m_n}),
\]
and hence
\[
2N_n
E_{\theta_0}
\left[
R(\widehat\theta_n^{(2)})-R^*
\mid
\mathcal F_{m_n}
\right]
=
\Phi(\widehat a_{m_n})+o_p(1).
\]

Theorem~\ref{thm:adaptive-oracle} gives
\[
\Phi(\widehat a_{m_n})
\overset{p}{\longrightarrow}
\Phi_\rho^\star,
\]
and therefore
\[
2N_n
E_{\theta_0}
\left[
R(\widehat\theta_n^{(2)})-R^*
\mid
\mathcal F_{m_n}
\right]
\overset{p}{\longrightarrow}
\Phi_\rho^\star.
\]

Since
\[
\frac{N_n}{n}
=
1-\frac{m_n}{n}
\longrightarrow1,
\]
the same first-order limit is obtained with \(N_n\) replaced by \(n\).

Finally, if the conditional scaled excess risks are uniformly integrable,
convergence in probability implies convergence of expectations. By the tower
property,
\[
2N_n
\left[
E_{\theta_0}
\{R(\widehat\theta_n^{(2)})\}
-
R^*
\right]
\longrightarrow
\Phi_\rho^\star.
\]

This completes the proof of Theorem~\ref{thm:adaptive-risk}.
\(\square\)
\section{Gaussian discriminant calculations}
\label{sec:supp-gaussian}

This section derives the information and boundary-curvature expressions used
in Section~\ref{sec:gaussian-specialization}.

\subsection*{Conditional label information}

Recall
\[
r_k(y;\theta)
=
\pi_k\phi_p(y;\mu_k,\Sigma_k),
\qquad
p_\theta(y)
=
\sum_{l=1}^{g}r_l(y;\theta),
\]
and define
\[
t_k(y;\theta)
=
\nabla_\theta\log r_k(y;\theta).
\]
Then
\[
\nabla_\theta r_k(y;\theta)
=
r_k(y;\theta)t_k(y;\theta).
\]
Consequently,
\begin{align}
\nabla_\theta\log p_\theta(y)
&=
\frac{
\sum_{k=1}^{g}\nabla_\theta r_k(y;\theta)
}{
p_\theta(y)
}
\nonumber\\
&=
\sum_{k=1}^{g}
\frac{r_k(y;\theta)}{p_\theta(y)}
t_k(y;\theta)
\nonumber\\
&=
\sum_{k=1}^{g}
\tau_k(y;\theta)t_k(y;\theta)
\nonumber\\
&=
\bar t(y;\theta).
\label{eq:supp-feature-score}
\end{align}

Since
\[
\log\tau_k(y;\theta)
=
\log r_k(y;\theta)-\log p_\theta(y),
\]
we obtain
\[
\nabla_\theta\log\tau_k(y;\theta)
=
t_k(y;\theta)-\bar t(y;\theta).
\]
Therefore
\begin{align}
J(y)
&=
E
\left[
S_{Z\mid Y}S_{Z\mid Y}^{\top}
\mid Y=y
\right]
\nonumber\\
&=
\sum_{k=1}^{g}
\tau_k(y)
\{t_k(y)-\bar t(y)\}
\{t_k(y)-\bar t(y)\}^{\top},
\end{align}
which proves \eqref{eq:gaussian-J}.

Because \(\bar t(Y)\) is the marginal feature score,
\[
I_Y
=
E\{\bar t(Y)\bar t(Y)^\top\}.
\]
Substitution into Proposition~\ref{prop:design-information} yields
\eqref{eq:gaussian-design-information}.

\subsection*{Explicit Gaussian class scores}

Under the baseline-logit parameterization,
\[
\pi_k
=
\frac{\exp(\alpha_k)}
{
1+\sum_{r=1}^{g-1}\exp(\alpha_r)
},
\qquad
k<g,
\]
and
\[
\pi_g
=
\frac{1}
{
1+\sum_{r=1}^{g-1}\exp(\alpha_r)
}.
\]
Hence, for \(j=1,\ldots,g-1\),
\[
\frac{\partial\log\pi_k}{\partial\alpha_j}
=
\begin{cases}
\mathbf 1(k=j)-\pi_j, & k<g,\\
-\pi_j, & k=g.
\end{cases}
\]
This gives \eqref{eq:prior-score}.

For the mean vector,
\[
\log\phi_p(y;\mu_k,\Sigma_k)
=
-\frac p2\log(2\pi)
-\frac12\log|\Sigma_k|
-\frac12
(y-\mu_k)^\top
\Sigma_k^{-1}
(y-\mu_k),
\]
and direct differentiation gives
\[
\nabla_{\mu_k}
\log r_k(y;\theta)
=
\Sigma_k^{-1}(y-\mu_k).
\]

For the covariance matrix, write
\[
q_k=y-\mu_k.
\]
The differential with respect to \(\Sigma_k\) is
\begin{align}
d\log r_k
&=
-\frac12
\operatorname{tr}
(\Sigma_k^{-1}d\Sigma_k)
+
\frac12
q_k^\top
\Sigma_k^{-1}
(d\Sigma_k)
\Sigma_k^{-1}
q_k
\nonumber\\
&=
\frac12
\operatorname{tr}
\left[
\Sigma_k^{-1}
\{q_kq_k^\top-\Sigma_k\}
\Sigma_k^{-1}
d\Sigma_k
\right].
\end{align}
Using
\[
\operatorname{vec}(d\Sigma_k)
=
D_p\,d\operatorname{vech}(\Sigma_k),
\]
we obtain
\[
\nabla_{\operatorname{vech}(\Sigma_k)}
\log r_k(y;\theta)
=
\frac12
D_p^\top
\operatorname{vec}
\left[
\Sigma_k^{-1}
\{q_kq_k^\top-\Sigma_k\}
\Sigma_k^{-1}
\right].
\]
This proves \eqref{eq:covariance-score}.

\subsection*{Proof of Proposition~\ref{prop:gaussian-face-curvature}}

For a pair \(k<l\), define
\[
g_{kl}(y;\theta)
=
r_k(y;\theta)-r_l(y;\theta).
\]
Its parameter derivative is
\begin{align}
\nabla_\theta g_{kl}(y;\theta)
&=
\nabla_\theta r_k(y;\theta)
-
\nabla_\theta r_l(y;\theta)
\nonumber\\
&=
r_k(y;\theta)t_k(y;\theta)
-
r_l(y;\theta)t_l(y;\theta).
\label{eq:supp-boundary-theta}
\end{align}

On the Bayes face
\[
\mathcal F_{kl}
=
\left\{
y:
r_k(y;\theta_0)=r_l(y;\theta_0)
>
r_r(y;\theta_0)
\text{ for }r\notin\{k,l\}
\right\},
\]
write
\[
r_{kl}(y)
=
r_k(y;\theta_0)
=
r_l(y;\theta_0).
\]
Then \eqref{eq:supp-boundary-theta} becomes
\[
b_{kl}(y)
=
r_{kl}(y)
\{t_k(y)-t_l(y)\}.
\]

For the feature derivative,
\[
\nabla_y r_k(y)
=
r_k(y)
\nabla_y\log r_k(y),
\]
and
\[
\nabla_y\log r_k(y)
=
-\Sigma_k^{-1}(y-\mu_k).
\]
Therefore, on \(\mathcal F_{kl}\),
\begin{align}
\nabla_y g_{kl}(y;\theta_0)
&=
-r_{kl}(y)
\Sigma_k^{-1}(y-\mu_k)
+
r_{kl}(y)
\Sigma_l^{-1}(y-\mu_l)
\nonumber\\
&=
r_{kl}(y)
\left[
\Sigma_l^{-1}(y-\mu_l)
-
\Sigma_k^{-1}(y-\mu_k)
\right]
\nonumber\\
&=
r_{kl}(y)d_{kl}(y).
\end{align}
Thus
\[
\|\nabla_y g_{kl}(y;\theta_0)\|
=
r_{kl}(y)\|d_{kl}(y)\|.
\]

The general face-curvature matrix is
\[
H_{kl}
=
\int_{\mathcal F_{kl}}
\frac{
b_{kl}(s)b_{kl}(s)^\top
}{
\|\nabla_y g_{kl}(s;\theta_0)\|
}
\,dS(s).
\]
Substituting the preceding expressions gives
\begin{align}
H_{kl}
&=
\int_{\mathcal F_{kl}}
\frac{
r_{kl}(s)^2
\delta t_{kl}(s)
\delta t_{kl}(s)^\top
}{
r_{kl}(s)\|d_{kl}(s)\|
}
\,dS(s)
\nonumber\\
&=
\int_{\mathcal F_{kl}}
r_{kl}(s)
\frac{
\delta t_{kl}(s)
\delta t_{kl}(s)^\top
}{
\|d_{kl}(s)\|
}
\,dS(s),
\end{align}
where
\[
\delta t_{kl}(s)
=
t_k(s)-t_l(s).
\]
This proves
\eqref{eq:gaussian-face-curvature}.
\(\square\)

\subsection*{Scalar representation of the acquisition value}

Let
\[
G_a=I(a)^{-1}H_RI(a)^{-1}
\]
and
\[
u_k(y)=t_k(y)-\bar t(y).
\]
Since
\[
J(y)
=
\sum_{k=1}^{g}
\tau_k(y)u_k(y)u_k(y)^\top,
\]
we have
\begin{align}
\psi_a(y)
&=
\operatorname{tr}\{G_aJ(y)\}
\nonumber\\
&=
\sum_{k=1}^{g}
\tau_k(y)
\operatorname{tr}
\{G_au_k(y)u_k(y)^\top\}
\nonumber\\
&=
\sum_{k=1}^{g}
\tau_k(y)
u_k(y)^\top G_au_k(y),
\end{align}
which proves
\eqref{eq:gaussian-value-score}.

Replacing \(H_R\) by \(H_{kl}\) gives
\[
G_{kl,a}
=
I(a)^{-1}H_{kl}I(a)^{-1}
\]
and
\[
\psi_{kl,a}(y)
=
\sum_{r=1}^{g}
\tau_r(y)
u_r(y)^\top
G_{kl,a}
u_r(y),
\]
which proves
\eqref{eq:gaussian-face-value-score}.
\section{Additional real-data application: Dry Bean data}
\label{sec:supp-drybean}

As an additional empirical application complementing the Landsat analysis
in Section~\ref{sec:realdata}, we considered the Dry Bean data set of
\citet{KokluOzkan2020}. The data
contain image-derived morphological measurements on seven varieties of dry
beans: Barbunya, Bombay, Cali, Dermason, Horoz, Seker, and Sira. The original
data set contains 13,611 observations and 16 quantitative predictors. After
removing 68 exact duplicate observations, 13,543 observations remained for
the analysis.

The original predictors exhibit substantial collinearity. Several pairs of
morphological measurements have correlations close to one, and direct
quadratic discriminant analysis in the original 16-dimensional space leads
to severely ill-conditioned class-specific covariance matrices. We therefore
standardized the predictors and retained the first five principal components,
which together explained approximately 97.8\% of the total predictor
variation. Within each replication, the standardization and
principal-component transformation were estimated from the training features
only and subsequently applied to the corresponding test sample.

The Gaussian QDA specification is used here as a working parametric
approximation rather than as an assumption that the Dry Bean population is
exactly generated by a Gaussian mixture. Accordingly, this application
examines the practical behavior of the acquisition criterion under possible
model misspecification rather than serving as a direct verification of the
oracle asymptotic theory.

We performed 50 repeated stratified train--test splits, allocating 70\%
of the observations to training and 30\% to testing. Each replication
therefore contained 9,478 training observations and 4,065 test observations.
All feature vectors in the training pool were regarded as observed, whereas
their class labels were treated as requiring acquisition. An initial random
pilot sample comprising 5\% of the training pool was labeled, corresponding
to 474 observations. A semi-supervised seven-class Gaussian discriminant
model was then fitted using the pilot labels together with the unlabeled
training features. Estimation used the semi-supervised Gaussian-mixture
likelihood via the EM algorithm, with the observed class memberships held
fixed and the remaining class memberships treated as latent. Posterior
class probabilities for the unlabeled observations were updated in the
E-step, and the class proportions, means, and class-specific covariance
matrices were updated in the M-step. A small eigenvalue floor was applied
to the covariance matrices for numerical stability.

We considered total labeling budgets of 10\%, 20\%, and 30\% of the
training pool. Including the common 5\% pilot sample, these correspond to
948, 1,896, and 2,844 labeled training observations, respectively. The
proposed adaptive classification-risk design was compared with random
acquisition, entropy sampling, margin sampling, and a Fisher-information
design. Within each replication, all methods used the same train--test split
and the same pilot sample, yielding paired comparisons between acquisition
rules.

For the Fisher and adaptive classification-risk rules, the relaxed
finite-pool design problem was solved by a Frank--Wolfe algorithm. Convergence
was assessed using the relative Frank--Wolfe optimality gap, with tolerance
$10^{-5}$, and the resulting relaxed design was converted to an exact
labeling set by deterministic top-budget rounding. The optimization was
stable across the experiment and no model-fitting failures occurred. After
the selected labels were revealed, the semi-supervised discriminant model
was refitted and evaluated on the held-out test observations.

\begin{figure}[H]
\centering
\includegraphics[width=\textwidth]{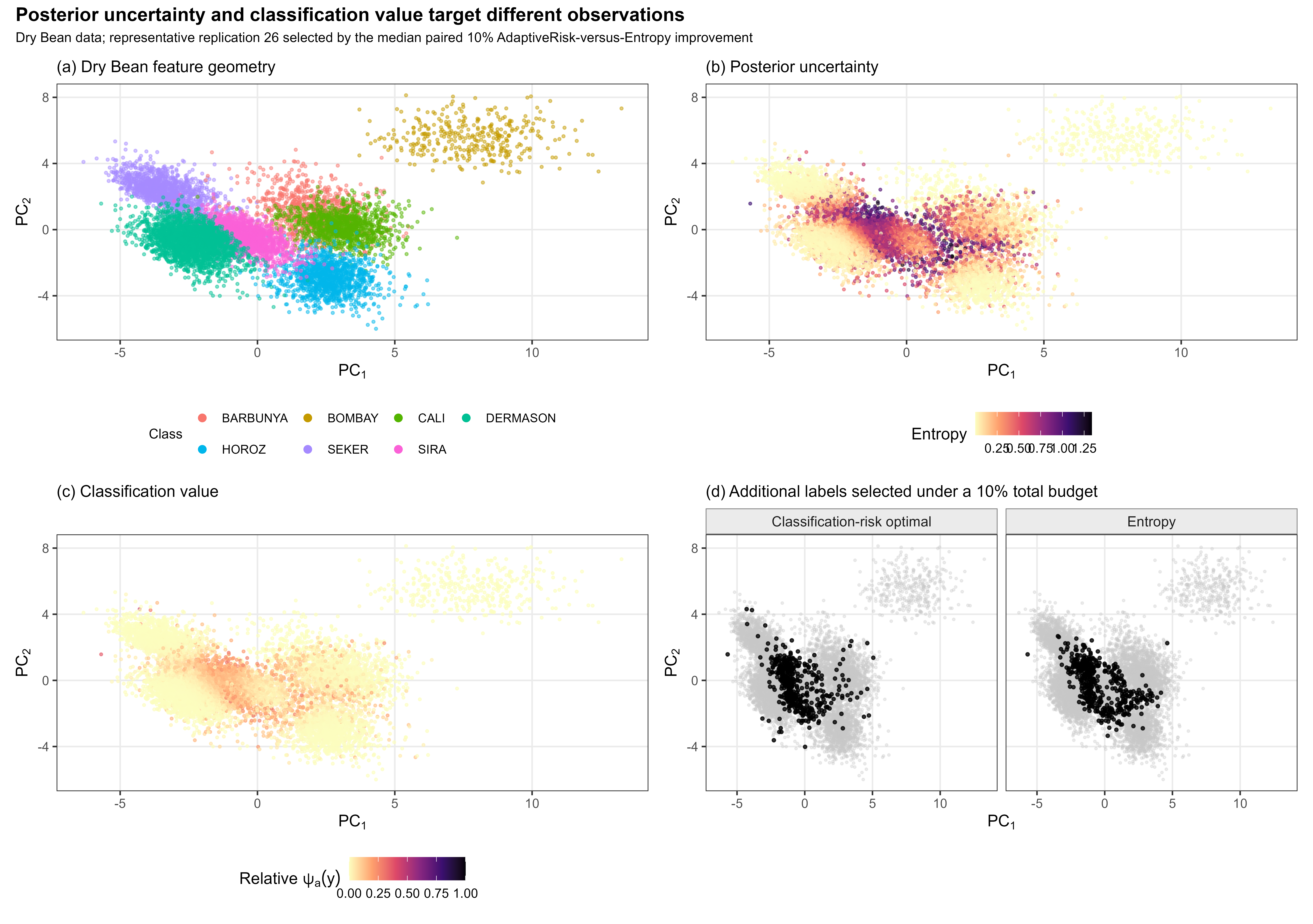}
\caption{
Acquisition geometry for the Dry Bean data in representative replication 26.
Panel (a) displays the seven observed classes in the first two principal
components. Panel (b) shows posterior classification uncertainty measured by
entropy, whereas panel (c) shows the estimated relative classification value
$\widehat{\psi}_a(y)$. Panel (d) compares the additional observations selected
by classification-risk-optimal and entropy acquisition under a 10\% total
labeling budget; black points denote acquired labels and gray points the
remaining training pool. The displayed principal components are used only
for visualization: posterior uncertainty, classification value, and
acquisition decisions are calculated from the full five-dimensional model.
The replication was selected as the one whose paired
AdaptiveRisk-versus-Entropy improvement was closest to the median paired
improvement across the 50 replications.
}
\label{fig:drybean-geometry}
\end{figure}

Figure~\ref{fig:drybean-geometry} illustrates the distinction between
posterior uncertainty and classification value on a representative
replication. To avoid selecting an unusually favorable example, the
replication was chosen objectively as the one whose paired improvement of
the adaptive rule over entropy sampling at the 10\% budget was closest to
the median paired improvement across the 50 replications. This criterion
selected replication 26. The first two principal components are used in
Figure~\ref{fig:drybean-geometry} only for visualization; posterior
probabilities, classification values, and acquisition decisions are all
computed from the full five-dimensional model.

The figure provides a direct empirical illustration of the distinction
between posterior uncertainty and classification value developed in
Section~\ref{sec:uncertainty}. Posterior uncertainty
is elevated throughout several overlapping regions of the feature space,
so entropy sampling concentrates labels broadly around observations for
which the fitted class probabilities are ambiguous. Classification value,
by contrast, is markedly nonuniform within these uncertain regions. The
adaptive criterion assigns greater value to observations whose conditional
label information is aligned with parameter directions that are important
for classification risk. Consequently, the two acquisition rules select
different subsets of the same unlabeled pool, as is visible in
Figure~\ref{fig:drybean-geometry}(d). Thus, high posterior uncertainty does
not by itself imply high classification value in this real-data example.

\begin{figure}[H]
\centering
\includegraphics[width=0.82\textwidth]{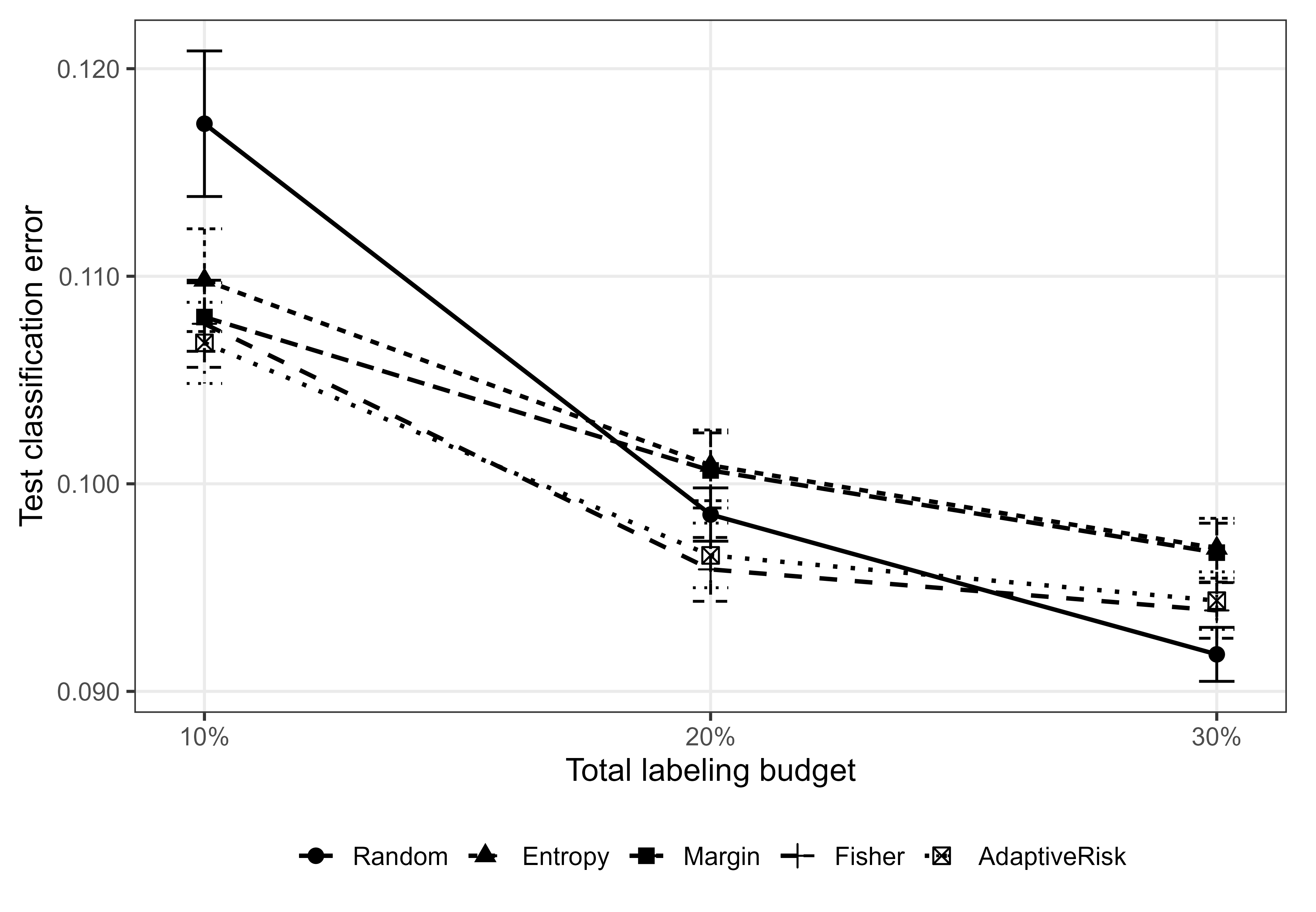}
\caption{
Mean test classification error for the Dry Bean data over 50 repeated
stratified train--test splits. Error bars represent $\pm1.96$ Monte Carlo
standard errors. Within each replication, all acquisition rules use the same
train--test split and initial pilot sample.
}
\label{fig:drybean-error}
\end{figure}

Table~\ref{tab:drybean-results} reports the mean overall and balanced
classification errors across the 50 replications. The corresponding mean
test errors and Monte Carlo uncertainty are displayed in
Figure~\ref{fig:drybean-error}.

\begin{table}[t]
\centering
\caption{
Mean test classification error and balanced error for the Dry Bean data over
50 repeated stratified train--test splits. Monte Carlo standard errors are
reported in parentheses. The smallest mean error within each labeling budget
is shown in bold.
}
\label{tab:drybean-results}

\resizebox{\textwidth}{!}{%
\begin{tabular}{lcccccc}
\hline
& \multicolumn{2}{c}{10\%}
& \multicolumn{2}{c}{20\%}
& \multicolumn{2}{c}{30\%} \\
\cline{2-3}\cline{4-5}\cline{6-7}
Method
& Error & Balanced
& Error & Balanced
& Error & Balanced \\
\hline
Random
& 0.11735 (0.00179)
& 0.09654 (0.00145)
& 0.09852 (0.00065)
& 0.08194 (0.00058)
& \textbf{0.09178} (0.00066)
& \textbf{0.07672} (0.00054)
\\
Entropy
& 0.10981 (0.00126)
& 0.09047 (0.00113)
& 0.10089 (0.00087)
& 0.08318 (0.00079)
& 0.09690 (0.00074)
& 0.08009 (0.00067)
\\
Margin
& 0.10803 (0.00084)
& 0.08864 (0.00081)
& 0.10064 (0.00092)
& 0.08308 (0.00085)
& 0.09668 (0.00072)
& 0.07994 (0.00067)
\\
Fisher
& 0.10771 (0.00107)
& 0.08940 (0.00094)
& \textbf{0.09588} (0.00078)
& \textbf{0.08015} (0.00073)
& 0.09390 (0.00068)
& 0.07823 (0.00066)
\\
Adaptive risk
& \textbf{0.10679} (0.00100)
& \textbf{0.08829} (0.00093)
& 0.09655 (0.00079)
& 0.08022 (0.00073)
& 0.09437 (0.00071)
& 0.07843 (0.00066)
\\
\hline
\end{tabular}%
}

\end{table}

At the smallest labeling budget, the adaptive classification-risk rule
attained the lowest mean test error, 0.1068, compared with 0.1077 for the
Fisher-information design, 0.1080 for margin sampling, 0.1098 for entropy
sampling, and 0.1173 for random acquisition. Relative to random acquisition,
this corresponds to an approximately 9\% reduction in mean
misclassification error. The adaptive rule also attained the smallest
balanced error, 0.0883, at this budget.

Because the methods were evaluated on common splits and pilot samples,
paired differences provide the most direct assessment of their relative
performance. At the 10\% labeling budget, the adaptive rule reduced mean
test error relative to random acquisition by 0.0106, with a 95\% Monte Carlo
interval of $(0.0074,0.0137)$, and relative to entropy sampling by 0.0030,
with a corresponding interval of $(0.0010,0.0050)$. Its improvements over
margin sampling and the Fisher-information design were smaller, approximately
0.0012 and 0.0009, respectively, and the corresponding intervals included
zero. The most pronounced empirical advantage of classification-risk
targeting at the smallest budget is therefore relative to random and
entropy-based acquisition.

At the 20\% labeling budget, the adaptive rule attained smaller mean test
error than random, entropy, and margin acquisition. The mean reductions in test error
were approximately 0.0020, 0.0043, and 0.0041, respectively. In particular, the paired 95\% Monte Carlo intervals for its improvements
over entropy and margin sampling were $(0.00366,0.00502)$ and
$(0.00338,0.00481)$. The
Fisher-information design attained the smallest mean error at this budget,
0.09588 compared with 0.09655 for the adaptive rule, a paired difference of
approximately $6.7\times10^{-4}$ in favor of Fisher information. Thus, as
the labeling budget increases, the empirical distinction between the two
information-based criteria becomes small.

At the 30\% labeling budget, the absolute differences between the methods
were again modest. Random acquisition attained the smallest mean error,
0.09178, followed by the Fisher and adaptive designs at 0.09390 and 0.09437,
respectively. The adaptive rule nevertheless continued to outperform entropy
and margin sampling, whose mean errors were 0.09690 and 0.09668. Its
difference from the Fisher-information design was only 0.00047 and was not
clearly distinguishable from zero in the paired comparison. The performance
of random acquisition at the largest budget also illustrates that the
finite-sample ordering of acquisition rules need not coincide uniformly with
the ordering of their local asymptotic design criteria.

The balanced-error results lead to essentially the same conclusions. The
adaptive rule produced the smallest balanced error at the 10\% budget,
whereas the Fisher and adaptive designs were nearly indistinguishable at
20\% and 30\%. Because the seven bean varieties have unequal class frequencies, the
corresponding improvement in balanced error indicates that the small-budget
result is not confined to the overall error criterion.

Overall, the Dry Bean analysis provides complementary empirical evidence for
the proposed framework. The observation-level visualization again shows that
posterior uncertainty and classification value can induce different rankings
of the same unlabeled feature pool. In addition, at the most restrictive
labeling budget, the adaptive classification-risk rule attained the smallest
mean overall and balanced classification errors among the five acquisition
rules. This advantage was not uniform across larger budgets: Fisher
information attained the smallest mean error at the 20\% budget, while
random acquisition attained the smallest mean error at 30\%. These results
therefore complement the Landsat application by illustrating both the
potential benefit of classification-risk targeting when labels are scarce
and the fact that the finite-sample ordering of acquisition rules need not
uniformly follow their local asymptotic design criteria.
%


\begin{thebibliography}{99}

\bibitem[Lewis and Gale, 1994]{lewis1994sequential}
Lewis, D. D. and Gale, W. A. (1994).
\newblock A sequential algorithm for training text classifiers.
\newblock In {\em Proceedings of the 17th Annual International ACM SIGIR Conference on Research and Development in Information Retrieval}, pages 3--12. Springer.
\newblock doi:10.1007/978-1-4471-2099-5\_1.

\bibitem[Lewis and Catlett, 1994]{lewis1994heterogeneous}
Lewis, D. D. and Catlett, J. (1994).
\newblock Heterogeneous uncertainty sampling for supervised learning.
\newblock In {\em Machine Learning Proceedings 1994}, pages 148--156. Morgan Kaufmann.
\newblock doi:10.1016/B978-1-55860-335-6.50026-X.

\bibitem[Settles, 2009]{settles2009active}
Settles, B. (2009).
\newblock Active learning literature survey.
\newblock Computer Sciences Technical Report 1648, University of Wisconsin--Madison.

\bibitem[Hoi et al., 2006]{hoi2006batch}
Hoi, S. C. H., Jin, R., Zhu, J., and Lyu, M. R. (2006).
\newblock Batch mode active learning and its application to medical image classification.
\newblock In {\em Proceedings of the 23rd International Conference on Machine Learning}, pages 417--424. ACM.
\newblock doi:10.1145/1143844.1143897.

\bibitem[Sourati et al., 2017]{sourati2017asymptotic}
Sourati, J., Akcakaya, M., Leen, T. K., Erdogmus, D., and Dy, J. G. (2017).
\newblock Asymptotic analysis of objectives based on Fisher information in active learning.
\newblock {\em Journal of Machine Learning Research}, 18(34):1--41.

\bibitem[Chen and Biros, 2023]{chen2023firal}
Chen, Y. and Biros, G. (2023).
\newblock {FIRAL}: An active learning algorithm for multinomial logistic regression.
\newblock In {\em Advances in Neural Information Processing Systems}, 36:65894--65905.
\newblock doi:10.52202/075280-2877.

\bibitem[Wang et al., 2018]{wang2018optimal}
Wang, H., Zhu, R., and Ma, P. (2018).
\newblock Optimal subsampling for large sample logistic regression.
\newblock {\em Journal of the American Statistical Association}, 113(522):829--844.
\newblock doi:10.1080/01621459.2017.1292914.

\bibitem[Yao and Wang, 2019]{yao2019optimal}
Yao, Y. and Wang, H. (2019).
\newblock Optimal subsampling for softmax regression.
\newblock {\em Statistical Papers}, 60(2):585--599.
\newblock doi:10.1007/s00362-018-01068-6.

\bibitem[Roy and McCallum, 2001]{roy2001toward}
Roy, N. and McCallum, A. (2001).
\newblock Toward optimal active learning through Monte Carlo estimation of error reduction.
\newblock In {\em Proceedings of the Eighteenth International Conference on Machine Learning}, pages 441--448. Morgan Kaufmann.

\bibitem[Mussmann et al., 2022]{mussmann2022active}
Mussmann, S., Reisler, J., Tsai, D., Mousavi, E., O'Brien, S., and Goldszmidt, M. (2022).
\newblock Active learning with expected error reduction.
\newblock {\em arXiv preprint arXiv:2211.09283}.
\newblock doi:10.48550/arXiv.2211.09283.

\bibitem[Filstroff et al., 2024]{filstroff2024targeted}
Filstroff, L., Sundin, I., Mikkola, P., Tiulpin, A., Kylm{\"a}oja, J., and Kaski, S. (2024).
\newblock Targeted active learning for Bayesian decision-making.
\newblock {\em Transactions on Machine Learning Research}.

\bibitem[Huang et al., 2026]{huang2026loss}
Huang, Z., Bickford Smith, F., and Rainforth, T. (2026).
\newblock Loss-driven Bayesian active learning.
\newblock In {\em Proceedings of the 29th International Conference on Artificial Intelligence and Statistics}, volume 300 of {\em Proceedings of Machine Learning Research}, pages 5140--5148. PMLR.

\bibitem[Wang et al., 2026]{wang2026learning}
Wang, Y.-G., Wu, J., and McLachlan, G. J. (2026).
\newblock Learning from uncertainty-dependent missing labels for semi-supervised classification.
\newblock {\em arXiv preprint arXiv:2608.23960}.
\newblock doi:10.48550/arXiv.2608.23960.

\bibitem[Setoudehtazangi and McLachlan, 2026]{SetoudehtazangiMcLachlan2026}
Setoudehtazangi, F. and McLachlan, G. J. (2026).
\newblock Informative label missingness in multiclass classification: Information geometry and excess risk.
\newblock {\em arXiv preprint arXiv:2608.30561}.
\newblock doi:10.48550/arXiv.2608.30561.

\bibitem[Koklu and Ozkan, 2020]{KokluOzkan2020}
Koklu, M. and Ozkan, I. A. (2020).
\newblock Multiclass classification of dry beans using computer vision and machine learning techniques.
\newblock {\em Computers and Electronics in Agriculture}, 174:105507.
\newblock doi:10.1016/j.compag.2020.105507.

\bibitem[Srinivasan, 1993]{Srinivasan1993}
Srinivasan, A. (1993).
\newblock Statlog (Landsat Satellite).
\newblock {\em UCI Machine Learning Repository}.
\newblock doi:10.24432/C55887.

\bibitem[Imberg et al., 2020]{imberg2020optimal}
Imberg, H., Jonasson, J., and Axelson-Fisk, M. (2020).
\newblock Optimal sampling in unbiased active learning.
\newblock In {\em Proceedings of the Twenty Third International Conference on Artificial Intelligence and Statistics}, volume 108 of {\em Proceedings of Machine Learning Research}, pages 559--569. PMLR.

\bibitem[Bach, 2006]{bach2006active}
Bach, F. R. (2006).
\newblock Active learning for misspecified generalized linear models.
\newblock In {\em Advances in Neural Information Processing Systems 19}, pages 65--72. MIT Press.

\bibitem[Zhang and Oles, 2000]{zhang2000probability}
Zhang, T. and Oles, F. J. (2000).
\newblock A probability analysis on the value of unlabeled data for classification problems.
\newblock In {\em Proceedings of the Seventeenth International Conference on Machine Learning}, pages 1191--1198. Morgan Kaufmann, San Francisco, CA.

\bibitem[Schein and Ungar, 2007]{schein2007active}
Schein, A. I. and Ungar, L. H. (2007).
\newblock Active learning for logistic regression: An evaluation.
\newblock {\em Machine Learning}, 68(3):235--265.
\newblock doi:10.1007/s10994-007-5019-5.

\bibitem[Bickford Smith et al., 2023]{bickfordsmith2023prediction}
Bickford Smith, F., Kirsch, A., Farquhar, S., Gal, Y., Foster, A., and Rainforth, T. (2023).
\newblock Prediction-oriented Bayesian active learning.
\newblock In {\em Proceedings of the 26th International Conference on Artificial Intelligence and Statistics}, volume 206 of {\em Proceedings of Machine Learning Research}, pages 7331--7348. PMLR.

\bibitem[Pukelsheim, 2006]{pukelsheim2006optimal}
Pukelsheim, F. (2006).
\newblock {\em Optimal Design of Experiments}.
\newblock Society for Industrial and Applied Mathematics, Philadelphia, PA.
\newblock doi:10.1137/1.9780898719109.

\bibitem[Hanneke and Yang, 2019]{hanneke2019surrogate}
Hanneke, S. and Yang, L. (2019).
\newblock Surrogate losses in passive and active learning.
\newblock {\em Electronic Journal of Statistics}, 13(2):4646--4708.
\newblock doi:10.1214/19-EJS1635.

\end{thebibliography}
\end{document}